\documentclass{article}
\usepackage[final,main]{neurips_2026}
\usepackage[utf8]{inputenc}
\usepackage[T1]{fontenc}
\usepackage{hyperref}
\usepackage{url}
\usepackage{amsfonts}
\usepackage{nicefrac}
\usepackage{microtype}
\usepackage{amsmath}
\usepackage{amssymb}
\usepackage{amsthm}
\usepackage{mathtools}
\usepackage{mathrsfs}
\usepackage{bm}
\usepackage{booktabs}
\usepackage{array}
\usepackage{tabularx}
\usepackage{colortbl}
\usepackage{xcolor}
\usepackage[normalem]{ulem}  
\usepackage{arydshln}   
\usepackage{multirow}
\usepackage{makecell}
\usepackage{graphicx}
\usepackage{caption}
\usepackage{subcaption}
\usepackage{float}
\usepackage{wrapfig}
\usepackage{tcolorbox}
\tcbuselibrary{skins, listings}
\usepackage{listings}
\usepackage{color}
\usepackage{enumitem}
\usepackage{pifont}
\usepackage{fontawesome5}
\usepackage{cancel}
\usepackage{balance}
\usepackage{placeins}
\usepackage{needspace}
\usepackage{algorithm}
\usepackage{algorithmic}

\definecolor{bestcolor}{RGB}{232,245,233}
\definecolor{secondcolor}{RGB}{255,243,224}
\definecolor{headercolor}{RGB}{63,81,181}
\definecolor{lightgray}{RGB}{245,245,245}

\theoremstyle{plain}

\theoremstyle{definition}

\theoremstyle{remark}

\definecolor{cardBlue}{HTML}{2878B5}
\definecolor{cardLightBlue}{HTML}{9AC9DB}
\definecolor{cardPeach}{HTML}{F8AC8C}
\definecolor{cardRed}{HTML}{C82423}
\definecolor{cardSalmon}{HTML}{FF8884}
 
\definecolor{cardBlueDark}{RGB}{28,68,130}
\definecolor{cardBlueSoft}{RGB}{232,241,250}
\definecolor{cardPeachSoft}{RGB}{254,243,237}
\definecolor{cardSalmonSoft}{RGB}{255,240,240}
\definecolor{cardRule}{RGB}{180,205,230}
\definecolor{cardGray}{RGB}{245,247,249}
\definecolor{cardSlate}{RGB}{80,100,130}
 
\newtcolorbox{cardbox}[2][]{%
  enhanced, 
  colback=cardBlueSoft,
  colframe=cardBlue,
  colbacktitle=cardBlue,
  coltitle=white,
  fonttitle=\bfseries\sffamily\small,
  title=#2,
  attach boxed title to top left={xshift=5mm, yshift=-2.5mm},
  boxed title style={
    size=small,
    colframe=cardBlueDark,
    sharp corners,
    colback=cardBlue
  },
  sharp corners=downhill,
  boxrule=0.65pt,
  left=3.5mm, right=3.5mm, top=3mm, bottom=3mm,
  #1
}
 
\newtcolorbox{cardnote}[1][]{%
  enhanced, 
  colback=cardPeachSoft,
  colframe=cardPeach!80!cardRed,
  sharp corners,
  boxrule=0.5pt,
  left=3mm, right=3mm, top=2mm, bottom=2mm,
  #1
}
 
\newtcolorbox{cardkey}[1][]{%
  enhanced, 
  colback=cardSalmonSoft,
  colframe=cardSalmon!90!cardRed,
  sharp corners,
  boxrule=0.5pt,
  left=3mm, right=3mm, top=2mm, bottom=2mm,
  #1
}
 
\definecolor{cardPromptBg}{RGB}{248,250,252}
\definecolor{cardPromptFrame}{RGB}{148,163,184}
\definecolor{cardPromptTitle}{RGB}{51,65,85}
\newtcolorbox{cardprompt}[2][]{%
  enhanced, 
  colback=cardPromptBg,
  colframe=cardPromptFrame,
  colbacktitle=cardPromptTitle,
  coltitle=white,
  fonttitle=\bfseries\sffamily\small,
  title=#2,
  attach boxed title to top left={xshift=5mm, yshift=-2.5mm},
  boxed title style={
    size=small,
    colframe=cardPromptTitle,
    sharp corners,
    colback=cardPromptTitle
  },
  sharp corners=downhill,
  boxrule=0.5pt,
  left=3.5mm, right=3.5mm, top=3mm, bottom=3mm,
  #1
}
 
\definecolor{cardWarnBg}{RGB}{255,241,241}
\definecolor{cardWarnFrame}{RGB}{185,28,28}
\definecolor{cardWarnTitle}{RGB}{153,27,27}
\newtcolorbox{cardwarn}[2][]{%
  enhanced, 
  colback=cardWarnBg,
  colframe=cardWarnFrame,
  colbacktitle=cardWarnTitle,
  coltitle=white,
  fonttitle=\bfseries\sffamily\small,
  title=#2,
  attach boxed title to top left={xshift=5mm, yshift=-2.5mm},
  boxed title style={
    size=small,
    colframe=cardWarnTitle,
    sharp corners,
    colback=cardWarnTitle
  },
  sharp corners=downhill,
  boxrule=0.5pt,
  left=3.5mm, right=3.5mm, top=3mm, bottom=3mm,
  #1
}

\definecolor{cardOrigBg}{RGB}{250,249,246}
\definecolor{cardOrigFrame}{RGB}{180,170,155}
\definecolor{cardOrigTitle}{RGB}{110,100,85}
\newtcolorbox{cardorig}[2][]{%
  enhanced, 
  colback=cardOrigBg,
  colframe=cardOrigFrame,
  colbacktitle=cardOrigTitle,
  coltitle=white,
  fonttitle=\bfseries\sffamily\small,
  title=#2,
  attach boxed title to top left={xshift=5mm, yshift=-2.5mm},
  boxed title style={
    size=small,
    colframe=cardOrigTitle,
    sharp corners,
    colback=cardOrigTitle
  },
  sharp corners=downhill,
  boxrule=0.5pt,
  left=3.5mm, right=3.5mm, top=3mm, bottom=3mm,
  #1
}

\lstdefinestyle{cardtree}{%
  basicstyle=\ttfamily\footnotesize,
  commentstyle=\color{cardSlate}\itshape,
  backgroundcolor=\color{cardBlueSoft},
  frame=none,
  showstringspaces=false,
  breaklines=true,
  columns=fullflexible,
  keepspaces=true,
  escapeinside={(*}{*)}
}
 
\lstdefinestyle{cardyaml}{%
  basicstyle=\ttfamily\footnotesize,
  keywordstyle=\color{cardBlue}\bfseries,
  commentstyle=\color{cardSlate}\itshape,
  stringstyle=\color{cardBlueDark},
  numbers=none,
  frame=single,
  rulecolor=\color{cardRule},
  backgroundcolor=\color{cardBlueSoft},
  showstringspaces=false,
  breaklines=true,
  morekeywords={model_path, vllm, tp, dp, gpu_memory_utilization,
    max_model_len, eval_batch_size, max_dedup_n, benchmarks,
    repeats, temperature, top_p, top_k, max_tokens,
    total_epochs, total_training_steps, save_freq, lora_rank,
    train_batch_size, max_response_length, strategy}
}

\title{From Solver Trajectories to Teaching Trajectories: Cognition-Aligned Reasoning Distillation}

\author{%
  \begin{minipage}{\dimexpr\textwidth-2\leftmargini\relax}
  \centering
  Yashuo Luo$^{1}$ \quad
  Qianren Mao$^{2,\dagger}$ \quad
  Ziqi Qin$^{3}$ \quad
  Junnan Liu$^{4}$ \quad
  Haitao Li$^{5}$ \\
  Qili Zhang$^{2,3}$ \quad
  Yuan Gao$^{1}$ \quad
  Zhijun Chen$^{6}$ \quad
  Xu Wang$^{3,\dagger}$ \\[0.5ex]
  \normalfont\fontsize{8.5pt}{10pt}\selectfont
  $^{1}$ National College for Excellent Engineers, Beihang University \quad
  $^{2}$ Zhongguancun Laboratory \\
  $^{4}$ Department of Data Science \& AI, Faculty of Information Technology, Monash University \\
  $^{3}$ Beihang University \quad $^{5}$ Tsinghua University \quad $^{6}$ Hong Kong Polytechnic University \\[0.5ex]
  luoyashuo@buaa.edu.cn; maoqr@zgclab.edu.cn; xuwang@buaa.edu.cn
  \end{minipage}
}

\begin{document}

\maketitle

\begingroup
\renewcommand{\thefootnote}{\fnsymbol{footnote}}
\begin{NoHyper}
\footnotetext[2]{Corresponding Authors.}
\end{NoHyper}
\endgroup

\begin{abstract}
Knowledge distillation from large reasoning models (LRMs) often uses raw teacher trajectories as supervision. However, this practice creates a mismatch between the goal of distillation and the nature of teacher trajectories. Advanced LRMs typically improve their reasoning capabilities through reward-driven post-training, such as RLHF. This training optimizes them as strong solvers, rather than as teachers that produce reasoning processes smaller models can easily learn. Since smaller models have different "cognitive capacities" compared to their larger teachers, directly imitating raw teacher trajectories can sometimes be ineffective and often requires a substantial number of high-quality samples. To bridge this mismatch, we propose \textbf{C}ognition-\textbf{A}ligned \textbf{R}easoning \textbf{D}istillation (\textbf{CARD}), an RL-based framework that transforms solver trajectories into teaching trajectories aligned with the student’s cognition. Avoiding expensive resampling, our framework directly transforms existing trajectories. We optimize CARD with two reward signals: (1) a Distribution Alignment Reward that matches the student’s predictive distribution, and (2) a Logical Coherence Reward that reduces misleading failed reasoning steps. Comprehensive evaluations across multiple student models and settings demonstrate that CARD outperforms distillation from raw teacher trajectories in in-domain settings while achieving substantial gains in out-of-distribution performance. {\hypersetup{hidelinks}\mbox{Code is available at \href{https://github.com/LiteSSLHub/CARD}{\faGithub\enspace\texttt{https://github.com/LiteSSLHub/CARD}}.}}
\end{abstract}

\section{Introduction}
\label{sec:introduction}

Large reasoning models~(LRMs), such as DeepSeek-R1~\cite{DBLP:journals/corr/abs-2501-12948},
OpenAI o1~\cite{openai2024o1}, Kimi K2.5~\cite{DBLP:journals/corr/abs-2602-02276},
and Gemini 3 Pro~\cite{googledeepmind2025gemini3proeval}, have demonstrated strong performance on complex reasoning tasks in mathematics, science, and code.
These models solve complex problems by generating chains of reasoning steps during inference, commonly known as chain-of-thought~(CoT) reasoning~\cite{DBLP:conf/nips/Wei0SBIXCLZ22}.
However, limited model capacity can make it difficult for small language models~(SLMs) to acquire such advanced reasoning behaviors through reinforcement learning alone~\cite{DBLP:journals/corr/abs-2501-12948}.~\footnote{We define small language models as decoder-only language models with typically fewer than 10B parameters in this paper.}
To bridge this gap, reasoning distillation transfers reasoning capability from large models to SLMs, and existing approaches have demonstrated competitive performance across reasoning
benchmarks~\cite{DBLP:journals/corr/abs-2402-03300,DBLP:journals/corr/abs-2501-12948,journals/corr/abs-2503-10460}.

\begin{figure}[t]
    \centering
    \includegraphics[width=\linewidth]{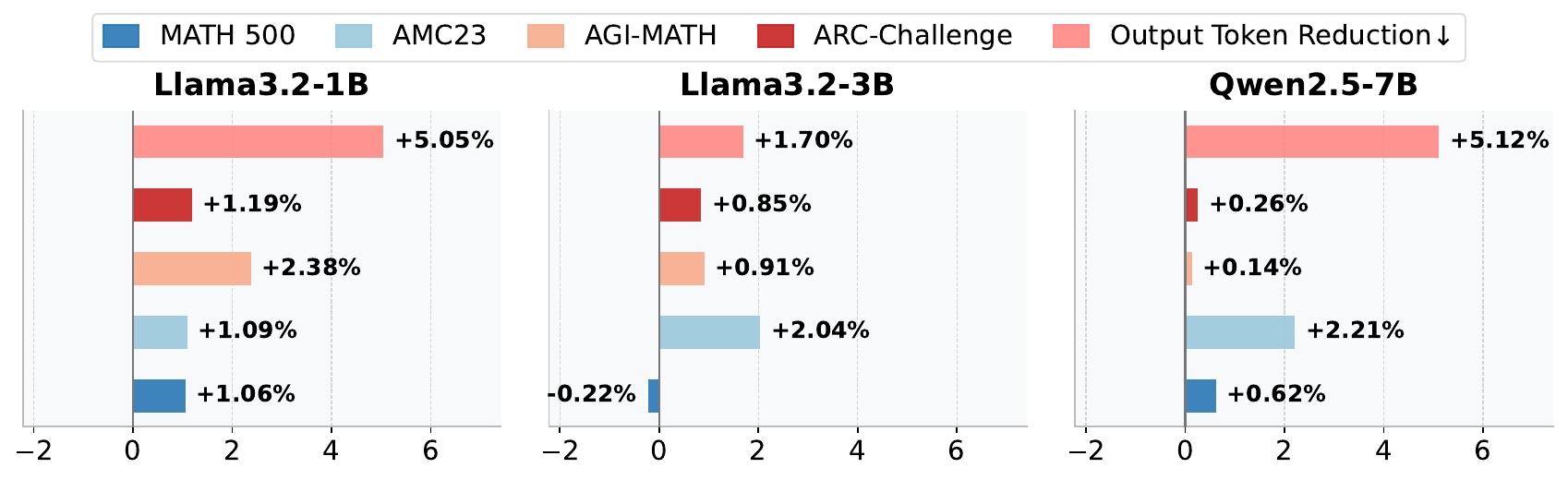}
    \caption{
Motivating analysis of \emph{failed steps} in \emph{trajectory-based} reasoning distillation.
Using DeepSeek-R1-generated DeepMath trajectories, we distill three student models
on two 10,000-trajectory subsets under identical settings, where one subset consists
entirely of trajectories with detected \emph{failed steps} and the other consists
entirely of trajectories without.
Bars report the difference between the no-failed-step subset and the failed-step subset.
Accuracy gains are reported in absolute points, while output-length reductions are reported as
$(\mathrm{Token}_{\mathrm{failed}}-\mathrm{Token}_{\mathrm{no\text{-}failed}})/\mathrm{Token}_{\mathrm{failed}}\times100\%$.
Evaluation includes math benchmarks and ARC-Challenge for out-of-distribution generalization,
and each panel corresponds to a different student model.
    }
    \label{fig:fsf_phenomenon}
\end{figure}

\begin{figure}[t]
    \centering
    \includegraphics[width=\linewidth, trim=0 0 0 0, clip]{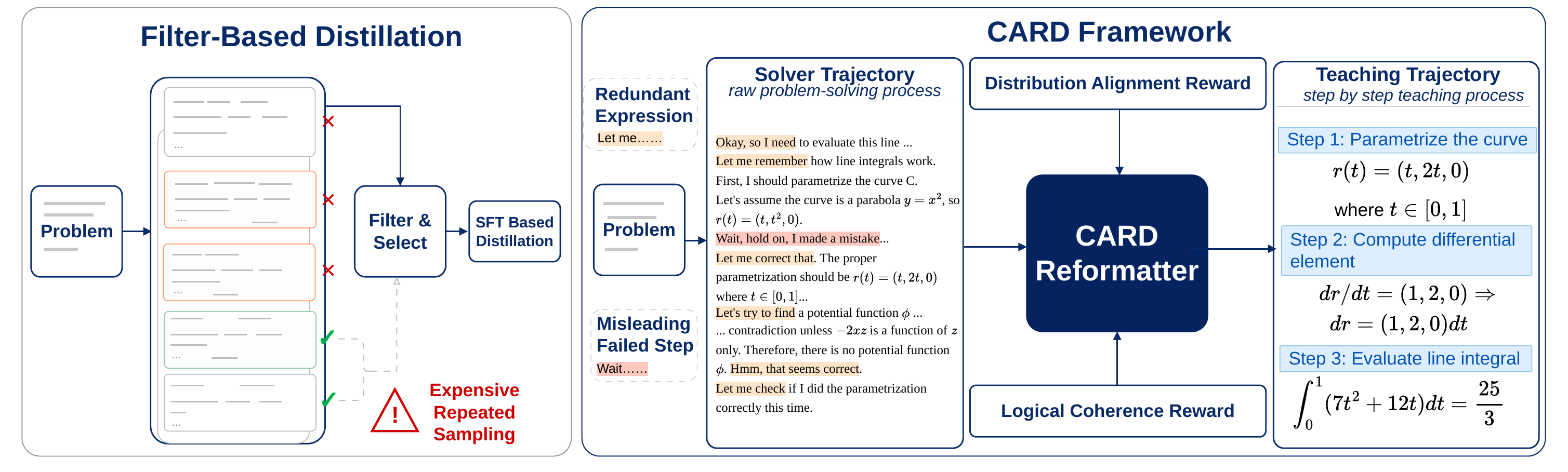}
    \vspace{-1.5em}
    \caption{%
Illustration of \emph{filter-based} distillation methods and CARD.
\textbf{Left:} \emph{filter-based} distillation methods sample multiple teacher trajectories
before selection and SFT, incurring expensive repeated teacher sampling.
\textbf{Right:} CARD selectively reformats solver trajectories into
cognition-aligned teaching trajectories without repeated teacher sampling.
Highlights in the solver trajectory indicate detected failed steps and redundant
expressions that hinder student learning.
    }
    \label{fig:card_illustration}
\vspace{-1.5em}
\end{figure}

Existing reasoning distillation approaches typically fall into two broad categories,
\emph{logits-based} and \emph{trajectory-based}.
\emph{Logits-based} methods~\cite{DBLP:conf/iclr/GuDLL24,DBLP:conf/icml/KoKKMPKY24,DBLP:conf/icml/KoCKDLZY25} leverage token-level probability distributions from the teacher,
either by supervising the student on teacher-generated trajectories at the logit level or by on-policy distillation where the student generates trajectories and the teacher provides logit-level supervision~\cite{lu2025onpolicy}.
These methods require access to the teacher's output logits and further assume
a shared vocabulary between teacher and student. 
\emph{Trajectory-based} methods instead train the student directly on teacher-generated \emph{rationales} via supervised fine-tuning~(SFT) or its variants~\cite{DBLP:conf/iclr/GuDLL24,DBLP:conf/icml/KoKKMPKY24,DBLP:journals/corr/abs-2502-12143}.~\footnote{We use \emph{rationale} to denote the CoT reasoning steps generated between a question and its final answer.}
To improve trajectory quality, \emph{filter-based} distillation methods sample multiple trajectories per question and select the best ones~\cite{DBLP:conf/emnlp/ChenWQWYZ23,DBLP:conf/emnlp/YanLPCXT25}.
As shown in Figure~\ref{fig:card_illustration}, this requires repeated teacher sampling and can be expensive for LRMs with large parameters.
More critically, filtering only selects among existing teacher-generated trajectories and does not adapt them to the student.
Beyond trajectory quality, direct SFT on task-specific trajectories
shifts the student toward the downstream distribution,
causing catastrophic forgetting of broader general capabilities~\cite{DBLP:conf/emnlp/KothaRBDB24,DBLP:conf/acl/LuoXZCL0ZL25}.

However, raw teacher-generated \emph{rationales} may not always serve as optimal supervision for student models.
Recent advanced LRMs primarily improve their reasoning capabilities through
reinforcement learning with verifiable rewards~\cite{DBLP:journals/corr/abs-2501-12948}, which optimizes them as strong problem solvers rather than as teachers whose trajectories SLMs can readily learn from.
We define the \emph{Cognition Gap} as the mismatch between what teacher-generated solver trajectories provide and what would better serve as supervision for student learning.
Prior work and our analysis suggest that the \emph{Cognition Gap} may surface in two ways: (1) Trajectories containing \emph{failed steps} reduce learning efficiency for SLMs.
As shown in Figure~\ref{fig:card_illustration}, teacher trajectories may include abandoned exploratory paths during reasoning, which we denote as
\emph{failed steps}~\cite{DBLP:journals/corr/abs-2509-19284}.
During teacher inference, \emph{failed steps} bias subsequent reasoning by leaving residual misleading context that the model does not fully discard after backtracking.
To assess how \emph{failed steps} in teacher trajectories affect student learning,
Figure~\ref{fig:fsf_phenomenon} compares SFT on trajectories with detected
\emph{failed steps} against SFT on trajectories without them under identical
training conditions.
Across both in-domain and out-of-domain~(OOD) benchmarks, the no-failed-step subsets generally outperform the failed-step subsets and also produce shorter average outputs, demonstrating that \emph{failed steps} degrade student learning efficiency in \emph{trajectory-based} knowledge distillation. (2) Even a correct teacher trajectory may be difficult for SLMs to learn from.
This relates to the Small Model Learnability Gap~\cite{DBLP:journals/corr/abs-2502-12143}, where small models do not consistently benefit from long CoT reasoning or distillation from larger teachers, and may instead learn more effectively from shorter, simpler reasoning chains that better match their capacity.

Intuitively, trajectories with \emph{rationales} that match the student's cognitive capacity should provide better teaching supervision than raw teacher-generated solver trajectories.
Based on this idea, we propose \textbf{CARD}~(\textbf{C}ognition-\textbf{A}ligned \textbf{R}easoning \textbf{D}istillation), a framework that reformats teacher-generated solver trajectories into cognition-aligned teaching trajectories without access to teacher model parameters.
As shown in Figure~\ref{fig:card_illustration}, a solver trajectory
records the teacher's problem-solving process and may include redundant
low-importance expressions or \emph{failed steps}. A teaching trajectory preserves the core reasoning process but presents it in clearer steps that the student can follow.

To convert teacher-generated solver trajectories into teaching trajectories,
CARD trains a lightweight Reformatter guided by two rewards.
(1) The \textbf{Distribution Alignment Reward} aligns each reformatted \emph{rationale} with the student's predictive distribution while preserving the answer behavior
induced by the original trajectory.
(2) The \textbf{Logical Coherence Reward} penalizes reformatted rationales that still contain detected \emph{failed steps}, thereby encouraging the Reformatter  to output coherent \emph{rationales}.
Unlike \emph{filter-based} distillation methods, CARD does not rely on repeated teacher sampling.
Instead, it leverages existing trajectories by selectively reformatting those
with detected \emph{failed steps} while keeping other trajectories unchanged.

Our major contributions are:
(1) We propose \textbf{CARD}, a \textbf{C}ognition-\textbf{A}ligned \textbf{R}easoning \textbf{D}istillation
framework that converts teacher-generated solver trajectories into teaching trajectories. 
CARD mitigates the \emph{Cognition Gap} by aligning reformatted \emph{rationales} with the student's predictive distribution and reducing misleading supervision from detected \emph{failed steps}.
(2) CARD offers a practical alternative to \emph{trajectory-based} knowledge distillation.
Without expensive repeated teacher sampling, CARD selectively reformats
existing solver trajectories containing detected \emph{failed steps}
while keeping other trajectories unchanged.
(3) We conduct extensive experiments across multiple student models, benchmarks, and evaluation settings.
CARD-reformatted teaching trajectories consistently outperform distillation from raw teacher-generated solver trajectories in in-domain settings while achieving substantial gains in OOD generalization.

\section{Related Work}
\label{sec:related_work}

\paragraph{Trajectory-Based Knowledge Distillation.}
Trajectory-based knowledge distillation trains the student on
teacher-generated \emph{rationales} via SFT~\cite{DBLP:conf/acl/HoSE23,
DBLP:conf/acl/LiHLJSL024}.
Filter-based methods improve the training corpus by selecting among existing
teacher-generated \emph{rationales}.
MCC-KD~\cite{DBLP:conf/emnlp/ChenWQWYZ23} samples multiple \emph{rationales}
per question and enforces consistency by minimizing the bidirectional KL
divergence between their answer distributions.
MoRSD~\cite{DBLP:conf/emnlp/YanLPCXT25} selects a compact subset by jointly
controlling accuracy, diversity, and \emph{rationale difficulty} via a student perplexity-based
Rationale Difficulty metric.
Beyond selection, \citet{DBLP:journals/corr/abs-2502-12143} find that SLMs
do not consistently benefit from distilling long CoT \emph{rationales} from
strong teachers, and fine-tuning on shorter, simpler chains can outperform
direct imitation of frontier LRMs.
Similarly, LS-Mixture SFT~\cite{DBLP:journals/corr/abs-2505-03469} pairs
original long CoT with structure-preserving short rewrites to mitigate
overthinking inherited from teacher models.
These methods generally operate on raw teacher-generated \emph{rationales}
without adapting their content to the student.

\paragraph{Quality of Chain-of-Thought.}
\citet{DBLP:journals/corr/abs-2501-19393} introduce budget forcing, which
appends a ``\texttt{Wait}'' token when the model would otherwise stop early,
to encourage longer reasoning and self-review, assuming that longer
\emph{rationales} are inherently beneficial.
TokenSkip~\cite{DBLP:conf/emnlp/XiaLWLL25} challenges this by showing that
long CoT \emph{rationales} contain redundant low-importance tokens, and trains
the model to skip them at inference time via token-level importance scores.
\citet{DBLP:journals/corr/abs-2509-19284} go further and find that the core
issue is not redundancy but \emph{failed steps}, defined as steps belonging
to abandoned exploratory branches in the CoT graph.
They define the \emph{Failed-Step Fraction} (FSF) as the proportion of
\emph{failed steps} in a trajectory, and causal experiments show that FSF
predicts reasoning accuracy more reliably than length or review ratio, and
that removing \emph{failed steps} improves subsequent reasoning accuracy.
\section{Preliminaries}
\label{sec:preliminaries}

We consider \emph{trajectory-based} reasoning distillation from a teacher LRM
$\mathcal{M}_T$ to a student SLM $\mathcal{M}_S(\theta)$.
Since $\mathcal{M}_T$ is accessible only through sampled outputs, distillation
is typically implemented via SFT or closely related variants.
Given an unlabeled problem set $\mathcal{P} = \{q_i\}_{i=1}^{N}$,
for each $q \in \mathcal{P}$ we draw a set of teacher-generated samples
\begin{equation}
      \mathcal{Y}_T(q) = \{y_j\}_{j=1}^{K}, \qquad y_j \sim \mathcal{M}_T(\cdot \mid q),
    \label{eq:define_sample}
\end{equation}
where each sample $y_j = (c_j, a_j)$ pairs a \emph{rationale} $c_j$ with an answer $a_j$.
A processing operator $\Phi$ constructs the distillation dataset
\begin{equation}
      \mathcal{D}_{\Phi}
        = \{(q,\,\Phi(q, \mathcal{Y}_T(q))) \mid q \in \mathcal{P}\},
    \label{eq:define_dataset}
\end{equation}
and the student is trained on $\mathcal{D}_{\Phi}$ via
\begin{equation}
      \mathcal{L}(\theta)
      = \mathbb{E}_{(q,y)\in\mathcal{D}_{\Phi}}
        \!\left[
          -\sum_{t=1}^{|y|} \log P_{\mathcal{M}_S}(y_t \mid q, y_{<t};\,\theta)
        \right].
    \label{eq:define_loss}
\end{equation}
Methods differ primarily in how $\Phi$ is designed.
Filter-based methods select a subset of teacher samples based on answer distributions
or difficulty metrics~\cite{DBLP:conf/emnlp/ChenWQWYZ23,DBLP:conf/emnlp/YanLPCXT25}.
\section{Method}
\label{sec:method}

\begin{figure}[t]
  \centering
  \includegraphics[width=\linewidth]{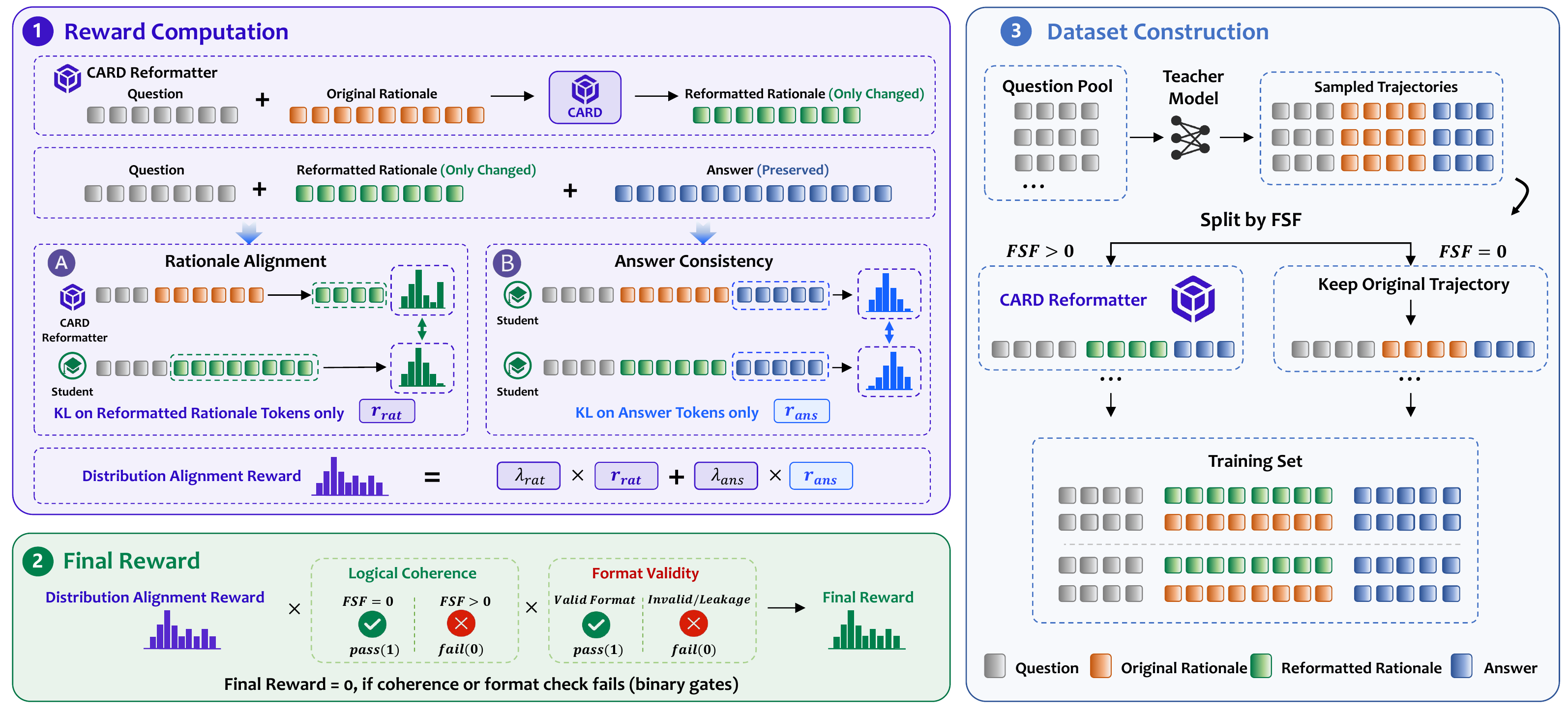}
\caption{Overview of CARD. \textbf{(1) Reward Computation.} The CARD Reformatter
takes a question and its original \emph{rationale} as input and produces a reformatted
\emph{rationale}, leaving the answer unchanged. Rationale Alignment ($r_{\mathrm{rat}}$)
and Answer Consistency ($r_{\mathrm{ans}}$) are combined into the Distribution
Alignment Reward via a weighted sum.
\textbf{(2) Final Reward.} The Distribution Alignment Reward is gated by a
Logical Coherence check (no \emph{failed steps}) and a Format Validity check (no
leakage); any candidate failing either gate receives a reward of zero.
\textbf{(3) Dataset Construction.} Teacher trajectories are split by their Failed-Step
Fraction (FSF): those with $\mathrm{FSF} > 0$ are reformatted by CARD, while
structurally clean trajectories with $\mathrm{FSF} = 0$ are retained directly
to form the final training set.}
\vspace{-1em}
  \label{fig:framework}
\end{figure}

CARD instantiates $\Phi$ (Eq.~\ref{eq:define_dataset}) as a \emph{rationale} 
Reformatter $\mathcal{R}_\phi$ that converts solver trajectories into 
cognition-aligned teaching trajectories better suited for student imitation.
Figure~\ref{fig:framework} provides an overview of the framework.
The Reformatter is optimized via a reward signal composed of two complementary 
terms, namely a Distribution Alignment Reward that measures how accessible the 
reformatted \emph{rationale} is to the student and a Logical Coherence Reward 
that penalizes reformatted \emph{rationales} containing detected \emph{failed steps}.
When constructing $\mathcal{D}_{\Phi}$ (Eq.~\ref{eq:define_dataset}), 
$\mathcal{R}_\phi$ is applied selectively so that only trajectories whose 
\emph{rationales} contain \emph{failed steps} are reformatted, while the 
remaining trajectories are retained directly.
The student model is then trained on $\mathcal{D}_{\Phi}$ via $\mathcal{L}(\theta)$ 
(Eq.~\ref{eq:define_loss}).
We describe each component in detail below.

\subsection{CARD Reformatter}

Even when a teacher \emph{rationale} leads to a correct answer, it may not be
suitable for student imitation.
As argued in Section~\ref{sec:introduction}, solver trajectories are optimized
for exploration and reward maximization, not for teaching.
CARD addresses this by training a \emph{rationale} Reformatter
$\mathcal{R}_\phi$ that maps each solver trajectory to a cognition-aligned \emph{rationale}
better suited for \emph{trajectory-based} knowledge distillation.

Specifically, given a training corpus
$\mathcal{D} = \{(q_i,\, c_i^{\mathrm{orig}},\, a_i)\}$
of (question, original \emph{rationale}, answer) triples,\footnote{For example, DeepSeek-R1~\cite{DBLP:journals/corr/abs-2501-12948} typically generates outputs of the form \texttt{<think>}$c$\texttt{</think>}$a$, where the enclosed content $c$ is the \emph{rationale} and the subsequent content $a$ is the answer.}
$\mathcal{R}_\phi$ takes a question $q$ and its original \emph{rationale}
$c^{\mathrm{orig}}$ as input and produces a cognition-aligned \emph{rationale}
$c^{\mathrm{ref}} = \mathcal{R}_\phi(q, c^{\mathrm{orig}})$,
while leaving the answer $a$ unchanged.
To preserve the teacher's final answer exactly and focus restructuring
where it matters most, we restrict reformatting to the \emph{rationale} $c$.
Complex reasoning is concentrated in the \emph{rationale} rather than the answer;
reformatting only this component retains the substantive instructional
content of the teacher trajectory while restructuring it for student accessibility.
The Reformatter is trained via a GRPO-based objective with the reward design
described below.

\subsection{Reward Design}

To achieve cognition alignment, we design two complementary rewards targeting
the two sources of mismatch identified in Section~\ref{sec:introduction}.
A \emph{Distribution Alignment Reward} targets the Cognition Gap by encouraging
cognition-aligned \emph{rationales} that fall within the student's cognitively
accessible range.
A \emph{Logical Coherence Reward} targets high Failed-Step Fraction by
rewarding reformatted \emph{rationales} that eliminate abandoned exploratory branches.

\paragraph{Distribution Alignment Reward.}
The goal of the \emph{rationale} Reformatter is not merely to paraphrase the teacher's
reasoning, but to restructure it so the student can follow the reasoning process
more easily.
We measure cognition alignment through two KL-based terms, both of which should
be minimized: lower values indicate that the reformatted \emph{rationale} is more
accessible to the student.

The first term, $r_{\mathrm{rat}}$, measures how much the student's
predictive distribution over the reformatted \emph{rationale} differs from the
Reformatter's distribution conditioned on the original solver trajectory:
\begin{equation}
  r_{\mathrm{rat}} =
  D_{\mathrm{KL}}\!\Bigl(
    \pi_\phi(c^{\mathrm{ref}}_t \mid q,\, c^{\mathrm{orig}},\, c^{\mathrm{ref}}_{<t})
    \;\Big\|\;
    \pi_{\mathrm{student}}(c^{\mathrm{ref}}_t \mid q,\, c^{\mathrm{ref}}_{<t})
  \Bigr).
  \label{eq:cot_kl}
\end{equation}
A small $r_{\mathrm{rat}}$ indicates that the student can anticipate the
cognition-aligned \emph{rationale} without the original solver trajectory,
reflecting genuine cognitive accessibility.
The second term, $r_{\mathrm{ans}}$, compares the student's answer
distributions conditioned on the cognition-aligned \emph{rationale} versus the original
solver trajectory:
\begin{equation}
  r_{\mathrm{ans}} =
  D_{\mathrm{KL}}\!\Bigl(
    \pi_{\mathrm{student}}(a_t \mid q,\, c^{\mathrm{ref}},\, a_{<t})
    \;\Big\|\;
    \pi_{\mathrm{student}}(a_t \mid q,\, c^{\mathrm{orig}},\, a_{<t})
  \Bigr).
  \label{eq:answer_kl}
\end{equation}
A small $r_{\mathrm{ans}}$ means that the answer distribution induced by
the cognition-aligned \emph{rationale} stays close to that induced by the original
solver trajectory, indicating that the reformatted \emph{rationale} remains semantically
coherent with the model's answer-generation process.
This avoids semantic discontinuity between the rewritten \emph{rationale} and the
answer portion of the original trajectory.

To mitigate reward instability caused by extreme values in the two KL terms, we normalize each term and apply a sigmoid before taking their weighted sum.
As both KL terms should be minimized, we negate them before normalization
so that higher reward corresponds to smaller divergence.
Concretely, let $\hat{r}_{\cdot,i} = -r_{\cdot,i}$ for each candidate $i$ in a
GRPO group of size $G$.
We apply group-wise z-score normalization:
\[
  \tilde{r}_{\cdot,i}
    = \frac{\hat{r}_{\cdot,i} - \hat{\mu}}{\hat{\sigma} + \epsilon},
  \qquad i \in \{1,\dots,G\},
\]
where $\hat{\mu}$ and $\hat{\sigma}$ are the mean and standard deviation of
$\hat{r}_{\cdot}$ within the group, and $\epsilon$ is a small constant for
numerical stability.
This normalization preserves the within-group ordering of candidates and improves
the stability of GRPO optimization~\cite{DBLP:journals/corr/abs-2501-12948}.
Each normalized term is then mapped to $(0,1)$ via sigmoid, yielding the
Distribution Alignment Reward for candidate $i$:
\begin{equation}
  r_{\mathrm{cog},i} =
    \lambda_{\mathrm{rat}} \cdot \sigma(\tilde{r}_{\mathrm{rat},i})
    + \lambda_{\mathrm{ans}} \cdot \sigma(\tilde{r}_{\mathrm{ans},i}).
  \label{eq:r_cognitive}
\end{equation}

\paragraph{Logical Coherence Reward.}
To mitigate the influence of misleading \emph{failed steps}, we introduce a reward
that encourages the Reformatter to produce structurally clean \emph{rationales}.
Following \citet{DBLP:journals/corr/abs-2509-19284}, we quantify structural
quality using the Failed-Step Fraction
$\mathrm{FSF}(c) \in [0,1]$, which measures the proportion of \emph{failed steps} in
$c$.
$\mathrm{FSF}(c)=0$ indicates a structurally clean reasoning chain free of
any detected \emph{failed steps}.
We compute FSF with an LLM-as-a-Judge pipeline, and defer the implementation
details to Appendix~\ref{sec:appendix_fsf}.
We define the Logical Coherence Reward as
\begin{equation}
  r_{\mathrm{coh},i}
  = \mathbb{I}\!\left[\mathrm{FSF}(c^{\mathrm{ref}}_i) = 0\right],
    \qquad i \in \{1,\dots,G\}, 
  \label{eq:r_coherence}
\end{equation}
which rewards cognition-aligned \emph{rationales} only when they contain no detected
\emph{failed steps}, thereby encouraging structurally continuous teaching trajectories.

\paragraph{Final Reward.}
During RL training, rollout outputs may exhibit reasoning information leakage,
where the Reformatter mentions modifying the original reasoning rather
than producing a clean cognition-aligned \emph{rationale}.
Appendix~\ref{sec:appendix_format_leakage} provides concrete examples.
We detect such leakage with lightweight rule-based pattern matching, detailed in
Appendix~\ref{sec:appendix_format_impl}.
Let $\mathbb{I}_{\mathrm{leak},i}$ denote whether candidate $i$ exhibits
reasoning information leakage.
The final reward is
\begin{equation}
R_i =
\bigl(1 - \mathbb{I}_{\mathrm{leak},i}\bigr)
\cdot r_{\mathrm{coh},i}
\cdot r_{\mathrm{cog},i},
\qquad i \in \{1,\dots,G\},
\label{eq:final_reward}
\end{equation}
where $r_{\mathrm{cog},i}$ measures how well the reformatted \emph{rationale}
aligns with the student's predictive distribution, while $r_{\mathrm{coh},i}$ and
$\mathbb{I}_{\mathrm{leak},i}$ act as binary gates: a \emph{rationale} containing
a detected \emph{failed step} or leakage receives zero reward.

\subsection{Reformatter Training and Dataset Construction}

We train the CARD Reformatter via GRPO using the reward defined in
Eq.~\ref{eq:final_reward}, optimizing $\mathcal{R}_\phi$ to produce
cognition-aligned \emph{rationales} that are both logically coherent and accessible
to the student.

Once training converges, we instantiate the processing operator $\Phi$
(defined in Eq.~\ref{eq:define_dataset}) using the trained CARD Reformatter.
Specifically, $\Phi$ applies $\mathcal{R}_\phi$ selectively: trajectories whose
\emph{rationales} satisfy $\mathrm{FSF} = 0$ are structurally clean and retained
directly, while those whose \emph{rationales} satisfy $\mathrm{FSF} > 0$ are
reformatted before inclusion.
This yields the distillation dataset $\mathcal{D}_\Phi$:
\begin{equation}
  \mathcal{D}_\Phi
  = \mathcal{D}^{\mathrm{FSF}=0}
    \;\cup\;
    \bigl\{(q_j,\;\mathcal{R}_\phi(q_j, c_j^{\mathrm{orig}}),\; a_j)
           \mid (q_j, c_j^{\mathrm{orig}}, a_j) \in \mathcal{D}^{\mathrm{FSF}>0}\bigr\},
  \label{eq:d2}
\end{equation}
where $\mathcal{D}^{\mathrm{FSF}=0}$ and $\mathcal{D}^{\mathrm{FSF}>0}$
denote the subsets of $\mathcal{D}$ whose \emph{rationales} have zero and
positive FSF, respectively.
The student model $\mathcal{M}_S$ is then trained on $\mathcal{D}_\Phi$ via the
standard SFT objective in Eq.~\ref{eq:define_loss}.
This selective strategy preserves structurally sound solver trajectories while
replacing fragmented ones with cognition-aligned \emph{rationales} that are
logically coherent and accessible to the student.
\section{Experiments}
\label{sec:experiment}

\definecolor{paletteblue}{HTML}{2878B5}
\definecolor{palettegreen}{HTML}{2E7D32}
\definecolor{palettered}{HTML}{C82423}
\definecolor{palettelightgreen}{HTML}{E6F4EA}
 
\colorlet{bestcolor}{palettelightgreen}        
\colorlet{basecolor}{black!6}                   
\colorlet{headercolor}{paletteblue!10}          
\colorlet{secondcolor}{white}        
\colorlet{improvecolor}{palettegreen}
\colorlet{dropcolor}{palettered!85!black}
 
\newcommand{\up}[1]{\textcolor{improvecolor}{\raisebox{0.2ex}{\tiny$\uparrow$#1}}}
\newcommand{\down}[1]{\textcolor{dropcolor}{\raisebox{0.2ex}{\tiny$\downarrow$#1}}}
 
\newcommand{\ccell}[2]{\multicolumn{1}{>{\columncolor{#1}[\tabcolsep][\tabcolsep]}c}{#2}}
\newcommand{\best}[1]{\ccell{bestcolor}{\textbf{#1}}}
\newcommand{\basecell}[1]{\ccell{basecolor}{#1}}
 
\newcommand{\second}[1]{\uline{#1}}
 
\begin{table*}[t]
\centering
\caption{
Main results on math and OOD generalization benchmarks.
\colorbox{bestcolor}{\textbf{Best}} results within each student block are highlighted.
$\uparrow$ denotes improvement of CARD over Vanilla SFT.
}
\label{tab:main}
\setlength{\tabcolsep}{3pt}
\renewcommand{\arraystretch}{1.15}
\footnotesize
\resizebox{\linewidth}{!}{
\begin{tabular}{@{}llccccccc c cc c@{}}
\toprule[1.5pt]
\multicolumn{1}{c}{\multirow{2}{*}{\textbf{Model}}} & \multicolumn{1}{c}{\multirow{2}{*}{\textbf{Method}}}
& \multicolumn{6}{c}{\textbf{Math}}
& \multicolumn{1}{c}{}
& \multicolumn{2}{c}{\textbf{General (OOD)}}
& \multicolumn{1}{c}{} \\
\cmidrule[0.8pt]{3-8} \cmidrule[0.8pt]{10-11}
&
& \textbf{AIME24}
& \textbf{AIME25}
& \textbf{GSM8K}
& \textbf{MATH500}
& \textbf{AMC23}
& \textbf{AGI-Math}
& \textbf{\textit{Avg}}
& \textbf{\shortstack{MMLU-Pro}}
& \textbf{ARC-C}
& \textbf{\textit{Avg}} \\
\midrule[1pt]
 
\multirow{7}{*}{\textbf{Qwen2.5-7B}}
& Base
& \basecell{10.42} & \basecell{5.83} & \basecell{90.17} & \basecell{73.80} & \basecell{41.30} & \basecell{70.60} & \basecell{48.69}
& \basecell{52.53} & \basecell{87.37} & \basecell{69.95} \\
\cmidrule{2-12}
& + Vanilla SFT
& \second{13.89} & \second{11.25} & 90.58 & 75.80 & 45.40 & \second{73.90} & \second{51.80}
& 45.60 & 67.07 & 56.34 \\
& + MCC-KD
& 12.50 & 9.50 & 90.10 & 74.70 & 44.00 & 72.50 & 50.55
& 46.04 & 67.41 & 56.73 \\
& + MoRSD
& 13.05 & 10.83 & 89.89 & 74.30 & 44.56 & 71.72 & 50.73
& 45.58 & 68.60 & 57.09 \\
& + Mix TokenSkip
& 12.50 & 10.00 & 88.62 & 68.60 & 42.80 & 69.90 & 48.74
& 44.37 & 71.16 & 57.77 \\
& + Mix Small Teacher
& 12.22 & 8.33 & \second{91.36} & \second{76.42} & \second{47.01} & 73.37 & 51.45
& \second{53.54} & \second{87.46} & \second{70.50} \\
\cmidrule{2-12}
& + CARD (ours)
& \best{16.39\up{2.50}} & \best{14.16\up{2.91}} & \best{91.87\up{1.29}} & \best{77.50\up{1.70}} & \best{48.50\up{3.10}} & \best{76.87\up{2.97}} & \best{54.22\up{2.42}}
& \best{54.00\up{8.40}} & \best{88.14\up{21.07}} & \best{71.07\up{14.73}} \\
\midrule[1.2pt]
 
\multirow{7}{*}{\textbf{Llama3.2-1B}}
& Base
& \basecell{0.14} & \basecell{0.00} & \basecell{26.91} & \basecell{14.40} & \basecell{4.07} & \basecell{17.03} & \basecell{10.43}
& \basecell{15.01} & \basecell{41.97} & \basecell{28.49} \\
\cmidrule{2-12}
& + Vanilla SFT
& 0.28 & 0.14 & 27.45 & 15.93 & 6.25 & \second{18.39} & 11.41
& 16.20 & 38.74 & 27.47 \\
& + MCC-KD
& 0.28 & 0.00 & 31.16 & \second{18.75} & \second{7.74} & 17.90 & \second{12.64}
& 15.15 & 36.06 & 25.61 \\
& + MoRSD
& \second{1.25} & \second{0.28} & 31.20 & 17.55 & 6.52 & 18.13 & 12.49
& 16.18 & 36.58 & 26.38 \\
& + Mix TokenSkip
& 0.28 & 0.00 & 30.23 & 17.02 & 7.61 & 18.18 & 12.22
& 15.68 & 33.85 & 24.77 \\
& + Mix Small Teacher
& 0.00 & 0.00 & \second{32.48} & 15.35 & 6.38 & 16.85 & 11.84
& \second{17.57} & \second{42.41} & \second{29.99} \\
\cmidrule{2-12}
& + CARD (ours)
& \best{1.38\up{1.10}} & \best{0.40\up{0.26}} & \best{33.97\up{6.52}} & \best{19.40\up{3.47}} & \best{8.97\up{2.72}} & \best{18.50\up{0.11}} & \best{13.77\up{2.36}}
& \best{18.88\up{2.68}} & \best{43.14\up{4.40}} & \best{31.01\up{3.54}} \\
\midrule[1.2pt]
 
\multirow{7}{*}{\textbf{Llama3.2-3B}}
& Base
& \basecell{0.28} & \basecell{0.83} & \basecell{63.08} & \basecell{31.03} & \basecell{10.87} & \basecell{35.40} & \basecell{23.58}
& \basecell{36.77} & \basecell{72.41} & \basecell{54.59} \\
\cmidrule{2-12}
& + Vanilla SFT
& 1.67 & 0.42 & 75.63 & 47.20 & 18.48 & 45.30 & 31.45
& \second{36.43} & 72.53 & 54.48 \\
& + MCC-KD
& 2.50 & \second{0.97} & 75.07 & 47.47 & \second{22.01} & 43.30 & 31.89
& 34.27 & 70.56 & 52.42 \\
& + MoRSD
& 2.35 & 0.83 & \second{76.42} & \second{47.70} & 21.74 & \second{45.40} & \second{32.41}
& 36.17 & 72.95 & 54.56 \\
& + Mix TokenSkip
& 1.94 & 0.14 & 73.84 & 40.38 & 17.26 & 42.62 & 29.36
& 36.37 & \second{73.18} & \second{54.78} \\
& + Mix Small Teacher
& \second{3.33} & 0.41 & 66.96 & 35.32 & 11.41 & 35.17 & 25.43
& 31.55 & 72.10 & 51.83 \\
\cmidrule{2-12}
& + CARD (ours)
& \best{3.74\up{2.07}} & \best{1.38\up{0.96}} & \best{77.26\up{1.63}} & \best{49.40\up{2.20}} & \best{23.13\up{4.65}} & \best{48.60\up{3.30}} & \best{33.92\up{2.47}}
& \best{36.91\up{0.48}} & \best{74.74\up{2.21}} & \best{55.83\up{1.35}} \\
 
\bottomrule[1.5pt]
\end{tabular}
}
\vspace{-1.5em}
\end{table*}

\subsection{Experimental Setup}
\label{sec:setup}

\paragraph{Training Data and Benchmarks}
\label{sec:exp_data}

DeepMath-103K provides three solver trajectories per question, 
generated by DeepSeek-R1-671B.
We partition the dataset into two disjoint splits:
5,000 questions to train the CARD Reformatter via GRPO,
and 30,000 questions for student SFT,
yielding 90,000 candidate trajectories (three per question).
Unless otherwise specified, each SFT method constructs a 
30,000-example training set from this pool.
We discard trajectories whose \emph{rationales} exceed 8,192 tokens
in either training stage (Reformatter training and student SFT).
We evaluate mathematical reasoning on six benchmarks: GSM8K, MATH-500, AGI-MATH, AMC 2023, AIME 2024, and AIME 2025, reporting average@8 for the first three and average@16 for the latter three to reduce variance on harder problems. The macro-average across all six serves as the combined math score. For out-of-domain~(OOD) generalization, we additionally evaluate on ARC-Challenge and MMLU-Pro (math questions excluded) with average@8. Further benchmark details appear in Appendix~\ref{sec:appendix_benchmarks}.

\paragraph{Models and Baselines}
\label{sec:exp_models}

We evaluate CARD on three student models: Llama-3.2-1B-Instruct,
Llama-3.2-3B-Instruct~\cite{DBLP:journals/corr/abs-2407-21783}, and
Qwen2.5-7B~\cite{DBLP:journals/corr/abs-2412-15115}, against
five baselines.
All distillation baselines use the same DeepSeek-R1-671B trajectory pool.
\textbf{Base} is the original instruct model; \textbf{Vanilla SFT} trains
directly on one trajectory randomly selected from each question's three candidates.
CARD uses the same training set as Vanilla SFT, applying the trained CARD
Reformatter to the $8{,}819$ trajectories with $\mathrm{FSF} > 0$.
The student is then fine-tuned on $\mathcal{D}_\Phi$ via Eq.~\ref{eq:define_loss}.
The \emph{filter-based} baselines are
MCC-KD~\cite{DBLP:conf/emnlp/ChenWQWYZ23} and MoRSD~\cite{DBLP:conf/emnlp/YanLPCXT25}.
MCC-KD selects 15{,}000 questions with two trajectories per question,
choosing among them by cross-\emph{rationale} answer consistency.
MoRSD selects 30{,}000 questions and picks one trajectory per question
from three candidates according to a \emph{rationale difficulty} metric
based on student perplexity.
To compare against mix strategies, we include Mix TokenSkip and Mix Small
Teacher, both sharing the same configuration as CARD and replacing the
\emph{rationales} of the $\mathrm{FSF} > 0$ subset.
Mix TokenSkip~\cite{DBLP:conf/emnlp/XiaLWLL25} replaces each such
\emph{rationale} with a TokenSkip rewrite.
Mix Small Teacher samples replacement trajectories from the Reformatter's
initial model.
Appendix~\ref{sec:appendix_baselines} provides implementation details for
all baselines.

\paragraph{Training Details}
\label{sec:exp_training}

Reformatter training uses rank-64, alpha-128 LoRA on Llama-3.1-8B-Instruct
and Qwen2.5-7B-Instruct for the Llama and Qwen students, respectively.
For student SFT, we fully fine-tune Llama3.2-1B/3B and use rank-128,
alpha-256 LoRA for Qwen2.5-7B.
Mix TokenSkip removes the 30\% least important \emph{rationale} tokens
using LLMLingua-2 (keep rate 0.7).
Full settings are in Appendix~\ref{sec:appendix_training}.


\Needspace{8\baselineskip}
\subsection{Main Results}
\label{sec:results}

Table~\ref{tab:main} compares CARD against Vanilla SFT and four other baselines
trained on the same DeepSeek-R1-671B trajectory pool.
Across all three student models, CARD achieves the strongest aggregate performance among the distillation methods, improving both in-domain~(ID) math performance and OOD generalization over Vanilla SFT.
\textbf{Compared with \emph{trajectory-based} distillation methods},
CARD improves Math Avg by $+2.36$/$+2.47$/$+2.42$ points on
Llama3.2-1B, Llama3.2-3B, and Qwen2.5-7B, respectively,
while improving General Avg by $+3.54$/$+1.35$/$+14.73$ points,
where the reported improvements are measured against the Vanilla SFT baseline.
The consistent ID gains suggest that reformatting \emph{rationales} into
cognition-aligned teaching trajectories provides stronger supervision
than passively selecting among raw teacher-generated trajectories,
effectively reducing the \emph{Cognition Gap} described in Section~\ref{sec:introduction}.
All three students also show consistent OOD improvements, with Qwen2.5-7B
achieving the most substantial gain of $+14.73$ points on General Avg.
This indicates that CARD's cognition-aligned \emph{rationales} effectively mitigate catastrophic forgetting induced by direct SFT on task-specific teacher trajectories~\cite{DBLP:conf/emnlp/KothaRBDB24,DBLP:conf/acl/LuoXZCL0ZL25}. In contrast, \emph{filter-based} methods suffer from this limitation because they directly leverage raw teacher-generated trajectories.
\textbf{Compared with Mix Small Teacher},
CARD outperforms it on Math Avg by $+1.93$/$+8.49$/$+2.77$ points,
suggesting the gains stem from the cognition-aligned training signal
rather than additional model capacity.
\textbf{Compared with Mix TokenSkip},
CARD outperforms it on Math Avg by $+1.55$/$+4.56$/$+5.48$ points,
indicating that cognition alignment requires restructuring the teacher's
core reasoning logic rather than merely compressing low-importance tokens.

\vspace{-0.5em}

\subsection{Comparison with FSF Filtering}
\label{sec:fsf_filtering}

\begin{wraptable}{r}{0.41\textwidth}
\vspace{-1.34em}
\centering
\footnotesize
\setlength{\tabcolsep}{4pt}
\captionsetup{justification=raggedright,singlelinecheck=false}
\caption{FSF filtering on student Qwen2.5-7B.}
\label{tab:fsf_filtering}
\renewcommand{\arraystretch}{1.18}
\begin{tabularx}{\linewidth}{l>{\centering\arraybackslash}X>{\centering\arraybackslash}X}
\toprule
\textbf{Method} & \textbf{Math Avg} & \textbf{Gen Avg} \\
\midrule
Vanilla SFT & 51.80 & 56.34 \\
Drop-FSF & 50.93 & 56.44 \\
Replace-FSF & 52.74 & 57.55 \\
\cellcolor{bestcolor}\textbf{CARD} & \cellcolor{bestcolor}\textbf{54.22} & \cellcolor{bestcolor}\textbf{71.07} \\
\bottomrule
\end{tabularx}
\vspace{-0.5em}
\end{wraptable}
We compare CARD with Drop-FSF, which discards trajectories with detected failed
steps, and Replace-FSF, which replaces them with trajectories containing no
detected failed steps at the original example budget. On Qwen2.5-7B,
CARD exceeds Replace-FSF by 1.48 Math Avg and 13.52 Gen Avg points
(Table~\ref{tab:fsf_filtering}), supporting a benefit beyond passive filtering.
Full per-benchmark scores are provided in Appendix~\ref{sec:appendix_fsf_controls}.

\par\WFclear

\subsection{Robustness across Random Seeds}
\label{sec:seed_robustness}

\begin{wraptable}{r}{0.41\textwidth}
\vspace{-1.34em}
\centering
\footnotesize
\setlength{\tabcolsep}{4pt}
\captionsetup{justification=raggedright,singlelinecheck=false}
\caption{Training-seed robustness on Qwen2.5-7B.}
\label{tab:seed_robustness}
\renewcommand{\arraystretch}{1.18}
\begin{tabularx}{\linewidth}{l>{\centering\arraybackslash}X>{\centering\arraybackslash}X}
\toprule
\textbf{Method} & \textbf{Math Avg} & \textbf{Gen Avg} \\
\midrule
Vanilla SFT & $52.34\pm0.64$ & $56.62\pm0.26$ \\
\cellcolor{bestcolor}\textbf{CARD} & \cellcolor{bestcolor}$\mathbf{54.94\pm0.69}$ & \cellcolor{bestcolor}$\mathbf{70.78\pm0.93}$ \\
\bottomrule
\end{tabularx}
\vspace{-0.5em}
\end{wraptable}
Table~\ref{tab:seed_robustness} reports mean $\pm$ sample standard deviation
across three training seeds on Qwen2.5-7B. CARD improves both aggregates in every
run, with paired gains of $2.60\pm0.22$ Math Avg and $14.16\pm0.94$ Gen Avg
points. These error bars capture training-run variability. Full per-seed results are provided in Appendix~\ref{sec:appendix_seed_evaluation}.
\par\WFclear

\subsection{Cross-Model Transferability of CARD}
\label{sec:card_transferability}
 
\begin{wraptable}{r}{0.41\textwidth}
\vspace{-1.3365em}
\centering
\footnotesize
\setlength{\tabcolsep}{4pt}
\captionsetup{justification=raggedright,singlelinecheck=false}
\caption{Cross-model transferability on student Llama3.2-3B.}
\label{tab:card_transferability}
\renewcommand{\arraystretch}{1.18}
\begin{tabularx}{\linewidth}{l>{\centering\arraybackslash}X>{\centering\arraybackslash}X}
\toprule
\textbf{Reward Student} & \textbf{Math Avg} & \textbf{Gen Avg} \\
\midrule
Llama3.2-1B & 32.26 & 55.13 \\
\cellcolor{bestcolor}\textbf{Llama3.2-3B} & \cellcolor{bestcolor}\textbf{33.92} & \cellcolor{bestcolor}\textbf{55.83} \\
Qwen2.5-7B & 33.02 & 55.24 \\
\bottomrule
\end{tabularx}
\vspace{-0.5em}
\end{wraptable}
 
To test transfer across reward students, we use Llama3.2-3B as the downstream
student and vary only the model used to compute rewards during Reformatter training.
As shown in Table~\ref{tab:card_transferability}, where \textit{Reward Student} denotes the student model used to compute
rewards during Reformatter training, the matched Llama3.2-3B
reformatter achieves the best performance across all benchmarks.
Notably, reformatters trained with mismatched students remain competitive.
The Qwen2.5-7B-trained reformatter reaches 33.02 Math Avg and 55.24 General Avg,
closing within 0.90 and 0.59 points of the matched Llama3.2-3B reformatter.
These results suggest that CARD learns \emph{rationale} transformations that generalize
across student models, rather than overfitting to a single reward student.
Full per-benchmark results are provided in Appendix~\ref{sec:appendix_ablation_full}.

\vspace{-0.5em}

\subsection{Analysis: Dynamics, Ablation and Robustness}
\begin{figure*}[h]
  \centering
  \begin{minipage}{0.93\linewidth}
  \centering
  \begin{subfigure}[t]{0.32\linewidth}
    \centering
    \includegraphics[width=\linewidth]{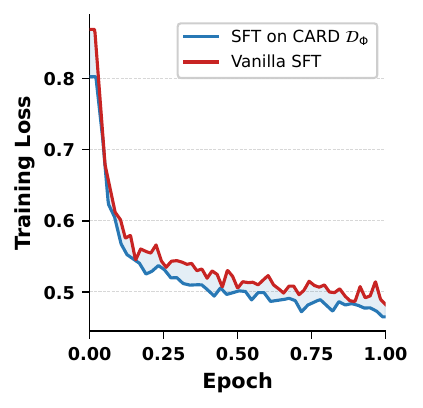}
  \end{subfigure}
  \hfill
  \begin{subfigure}[t]{0.32\linewidth}
    \centering
    \includegraphics[width=\linewidth]{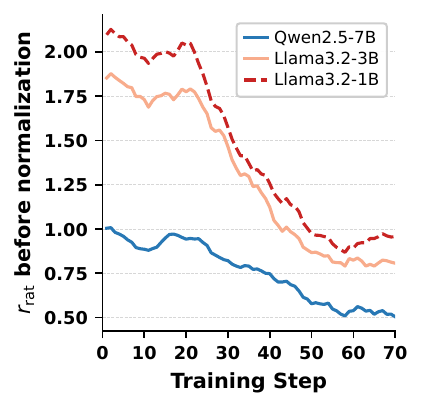}
  \end{subfigure}
  \hfill
  \begin{subfigure}[t]{0.32\linewidth}
    \centering
    \includegraphics[width=\linewidth]{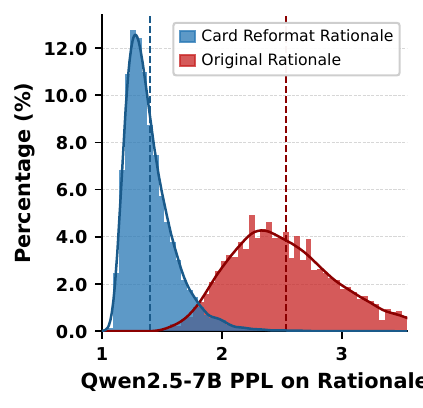}
  \end{subfigure}
  \end{minipage}
  \caption{Training dynamics of CARD. \textbf{Left}: SFT loss on Llama3.2-3B. \textbf{Middle}: $r_{\mathrm{rat}}$ during GRPO training. \textbf{Right}: PPL distribution.}
  \vspace{-1.0em}
  \label{fig:training_dynamics}
\end{figure*}

\paragraph{Training Dynamics}
\label{sec:training_dynamics}
We analyze training dynamics to verify that CARD drives the reformatted rationales
closer to the student distribution.
As shown in Figure~\ref{fig:training_dynamics}, CARD-reformatted rationales yield lower SFT loss than Vanilla SFT
on student Llama3.2-3B, suggesting that the reformatted rationales provide a more learnable supervised signal.
Moreover, the steady decrease of $r_{\mathrm{rat}}$ during GRPO training further supports that the Reformatter
is effectively optimized toward the KL-based Distribution Alignment Reward.
To directly quantify this distributional shift, we compute the PPL of Qwen2.5-7B on 5{,}000 trajectories
before and after CARD reformatting. We find that the PPL distribution shifts toward lower values after reformatting,
confirming that CARD brings the training data closer to the student distribution.
\vspace{-1.0em}

\begin{figure*}[h]
  \centering
  \begin{minipage}{0.93\linewidth}
  \centering
  \begin{subfigure}[t]{0.32\linewidth}
    \centering
    \includegraphics[width=\linewidth]{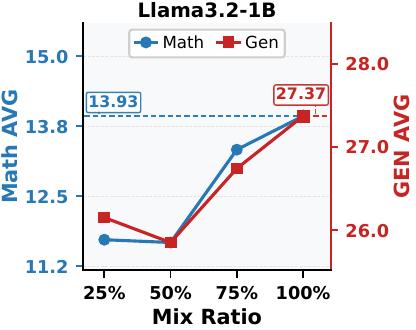}
  \end{subfigure}
  \hfill
  \begin{subfigure}[t]{0.32\linewidth}
    \centering
    \includegraphics[width=\linewidth]{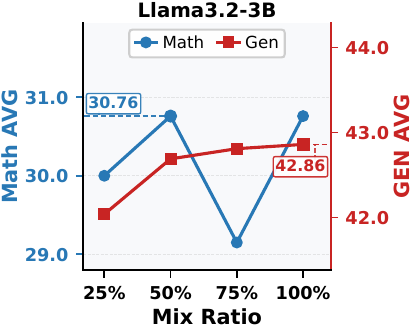}
  \end{subfigure}
  \hfill
  \begin{subfigure}[t]{0.32\linewidth}
    \centering
    \includegraphics[width=\linewidth]{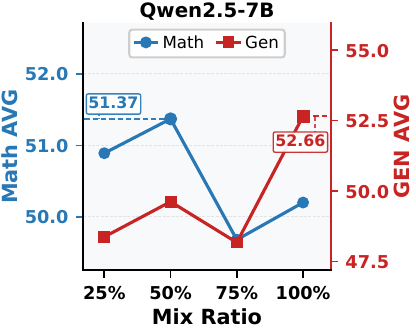}
  \end{subfigure}
  \end{minipage}
  \caption{Effect of reformatting selection criterion on student performance
under random selection across different fractions of the full training set.
Peak Math Avg and Gen Avg are annotated.}
  \vspace{-1em}
  \label{fig:mix_ratio}
\end{figure*}

\paragraph{Ablation on Reformatting Selection Criterion}
\label{sec:ablation_mix_ratio}
CARD applies $\mathcal{R}_\phi$ via FSF-guided selection, where only
trajectories in $\mathcal{D}^{\mathrm{FSF}>0}$ are reformatted while
$\mathcal{D}^{\mathrm{FSF}=0}$ trajectories are retained directly.
To investigate the effect of replacing it with random selection, we
vary the proportion of randomly selected trajectories from the full
training set submitted to $\mathcal{R}_\phi$: 25\%/50\%/75\%/100\%.
As shown in Figure~\ref{fig:mix_ratio}, random reformatting generally
degrades OOD generalization, with Gen Avg peaking at only
27.37/42.86/52.66 for Llama3.2-1B/Llama3.2-3B/Qwen2.5-7B,
compared with 31.01/55.83/71.07 under FSF-guided selection.
ID performance follows the same trend for Llama3.2-3B and Qwen2.5-7B (30.76 vs.\ 33.92 and 51.37 vs.\ 54.22), while the Math Avg gap for Llama3.2-1B remains marginal, suggesting FSF-guided selection provides a more consistent advantage with larger models.

\Needspace{14\baselineskip}
\paragraph{Ablation on CARD Components}
\label{sec:ablation_components}
\providecommand{\abest}[1]{\cellcolor{bestcolor}\textbf{#1}}
\providecommand{\asecond}[1]{\cellcolor{secondcolor}\uline{#1}}
\begin{wraptable}{r}{0.36\linewidth}
\vspace{-0.1365em}
\centering
\footnotesize
\setlength{\tabcolsep}{4pt}
\captionsetup{justification=raggedright,singlelinecheck=false}
\caption{CARD component ablations on Qwen2.5-7B.}

\begin{tabular}{lcc}
\toprule
\textbf{Method} & \textbf{Math Avg} & \textbf{Gen Avg} \\
\midrule
CARD                              & \abest{54.22}                & \abest{71.07} \\
w/o $r_{\mathrm{coh}}$            & 52.54\,\down{1.68}           & 66.31\,\down{4.76} \\
w/o $r_{\mathrm{rat}}$            & 52.23\,\down{1.99}           & 68.32\,\down{2.75} \\
w/o $r_{\mathrm{ans}}$            & 52.07\,\down{2.15}           & 68.63\,\down{2.44} \\
w/o $\mathbb{I}_{\mathrm{leak}}$  & \asecond{52.66}\,\down{1.56} & \asecond{69.17}\,\down{1.90} \\
\bottomrule
\end{tabular}
\vspace{-1em}
\end{wraptable}
We ablate the reward terms in Eq.~\ref{eq:final_reward}
on Qwen2.5-7B; full per-benchmark results are in
Table~\ref{tab:ablation_components_full}.
Removing any component degrades both ID and OOD performance, confirming
that every component is necessary.
For ID performance, removing $r_{\mathrm{ans}}$ causes the largest Math Avg
drop (54.22  vs.\  52.07), suggesting that preserving the answer behavior
induced by the original trajectory is most critical for in-domain tasks.
For OOD generalization, removing $r_{\mathrm{coh}}$ causes the largest Gen Avg
drop (71.07  vs.\  66.31), indicating that penalizing \emph{failed steps} in
reformatted \emph{rationales} is most critical for generalization.
Removing $r_{\mathrm{rat}}$ or $\mathbb{I}_{\mathrm{leak}}$ also degrades
performance, confirming that student-accessible \emph{rationales} and
leakage-free outputs both contribute to effective distillation.

\par\WFclear
\paragraph{Robustness to FSF Judge}
\label{sec:fsf_judge_robustness}
\begin{wraptable}{r}{0.36\linewidth}
\vspace{-1.3365em}
\centering
\footnotesize
\setlength{\tabcolsep}{4pt}
\captionsetup{justification=raggedright,singlelinecheck=false}
\caption{Robustness to FSF judge bias on student Qwen2.5-7B.}
\label{tab:fsf_judge_robustness}
\begin{tabular}{lcc}
\toprule
\textbf{Method} & \textbf{Math Avg} & \textbf{Gen Avg} \\
\midrule
CARD                       & \abest{54.22}                 & \abest{71.07} \\
w/ 5\% bias          & 52.88\,\down{1.34}            & 68.70\,\down{2.37} \\
w/ 10\% bias         & \asecond{53.37}\,\down{0.85}  & \asecond{68.94}\,\down{2.13} \\
w/ 20\% bias         & 52.91\,\down{1.31}            & 66.42\,\down{4.65} \\
\bottomrule
\end{tabular}
\vspace{-0em}
\end{wraptable}

CARD relies on the FSF judge to provide correctness signals at two points:
the Logical Coherence Reward (Eq.~\ref{eq:r_coherence}) and the selective
dataset construction (Eq.~\ref{eq:d2}), both of which hinge on the binary
judgment $\mathbb{I}[\mathrm{FSF}(\cdot) = 0]$.
To assess robustness to potential biases in this judge, we simulate
degraded judgment by independently flipping each binary FSF decision
with probability $p \in \{5\%, 10\%, 20\%\}$ at both locations above,
and evaluate on Qwen2.5-7B.
Full per-benchmark results appear in Table~\ref{tab:fsf_judge_robustness_full}.
As shown in Table~\ref{tab:fsf_judge_robustness}, injecting FSF judge bias
consistently degrades both ID and OOD performance, yet Math Avg remains
within $1.5$ points of the unperturbed baseline even under $20\%$ random
flipping ($52.91$ vs.\ $54.22$).
The degradation is markedly larger on OOD benchmarks (Gen Avg),
ranging from $2.13$ to $4.65$ points across bias levels, consistent with
our finding that cognition-aligned rationales are particularly important
for OOD generalization.

\section{Conclusion and Limitations}
\label{sec:conclusion}
\label{sec:limitations}

\textbf{Conclusion.}
In this paper, we presented \textbf{Cognition-Aligned Reasoning Distillation (CARD)}, which addresses the \emph{Cognition Gap} in trajectory-based knowledge distillation by reformatting raw solver trajectories into teaching trajectories better aligned with student cognition. The GRPO-trained Reformatter is guided by a Distribution Alignment Reward that matches reformatted rationales to the student's predictive distribution, and a Logical Coherence Reward that reduces misleading supervision from failed steps. Across three student models and eight benchmarks, CARD consistently outperforms distillation from raw solver trajectories on in-domain mathematical reasoning, while achieving even larger gains in out-of-domain generalization, where cognition-aligned rationales mitigate catastrophic forgetting induced by task-specific SFT.

\textbf{Limitations.}
Our evaluation focuses on mathematical reasoning with student models up to 7B
parameters. Due to limited computational resources, we have not evaluated CARD
with larger students or on multi-turn agentic tasks. Studying its scaling
behavior and extending cognition-aligned reasoning distillation to these
interactive settings remain directions for future work.

\begingroup
\renewcommand{\acksection}{\section*{Acknowledgments}}
\begin{ack}
This work was supported by Zhongguancun Laboratory and the National Cyber Security-National Science and Technology Major Project (Grant No. 2025ZD1503000). We thank the anonymous reviewers and the area chair for their careful reading and constructive feedback. Their detailed comments helped us clarify the methodological explanations, strengthen the experimental analysis, and improve the presentation of this work. We are grateful for their suggestions on the reliability and interpretation of the results.
\end{ack}
\endgroup

\bibliographystyle{abbrvnat}
\bibliography{references}

\newpage

\appendix


\section{Implementation Details}
\label{sec:appendix_impl}

\subsection{Rationale Reformatter}
\label{sec:appendix_reformatter}

The rationale reformatter $\mathcal{R}_\phi$ is trained on rollouts produced
by the following prompt template.
The system prompt defines the reformatting task and its constraints; the user
prompt supplies the concrete question and solver trajectory for each instance.

\begin{cardprompt}[label={box:system_prompt}]{System Prompt \normalfont\small for the Rationale Reformatter}
\small\ttfamily
You are an expert in math and code. The 'Original Reasoning Process'
provides a raw derivation that leads to a correct solution. Your task
is to extract the correct path and present it clearly.

\medskip
Follow these guidelines:

\medskip
1.~Refine and Bridge: Retain the core logic, including essential
   self-reflections and verification steps. Supplement logical leaps
   and clarify complex steps to ensure a smooth, coherent flow.

\medskip
2.~Filter Invalid Paths: Remove derivation branches within the
   `Original Reasoning' that do not lead to the final answer (e.g.,
   failed attempts, incorrect directions, or dead ends). Keep only
   the effective reasoning path.

\medskip
3.~Avoid Literal Repetition: Rephrase the content in your own words
   rather than simply duplicating the text of the original process.

\medskip
Output Rules:

- Directly output the rewritten reasoning process.

- Do NOT mention that you are rewriting another reasoning process.
\end{cardprompt}

\begin{cardprompt}[label={box:user_prompt}]{User Prompt Template \normalfont\small for the Rationale Reformatter}
\small\ttfamily
\#\# Problem

\{question\}

\medskip
\#\# Original Reasoning Process

\{thought\}

\medskip
\#\# Rewritten Reasoning Process
\end{cardprompt}

\subsection{Reward Design}
\label{sec:appendix_reward}

\subsubsection{Reasoning Information Leakage}
\label{sec:appendix_format_leakage}

During RL training of the CARD Reformatter, we observe that the reformatter occasionally produces outputs that reveal an awareness of the reformatting task, rather than presenting clean teaching trajectories.
We refer to this failure mode as \emph{reasoning information leakage} and mitigate it using the format penalty described in Section~\ref{sec:appendix_format_impl}.
Representative examples are shown below.

\begin{cardwarn}[label={box:leakage_patterns}]{Reasoning Information Leakage}
\small\ttfamily
\textit{Let's \textcolor{red!70!black}{refine and bridge the original reasoning process} to provide a clear path to the solution \ldots}

\medskip
\textit{\textcolor{red!70!black}{\#\# Problem}~~Let X and Y have the joint probability density function \ldots}

\medskip
\textit{\ldots This result is \textcolor{red!70!black}{consistent with the original reasoning process}, and it provides a clear and concise derivation.}

\medskip
\textit{We start by \textcolor{red!70!black}{rewriting} the product of square roots as a product of powers of 2 \ldots}
\end{cardwarn}

\subsubsection{Format Penalty}
\label{sec:appendix_format_impl}

To suppress reasoning information leakage, we apply a lightweight binary format
check to each rollout output before reward computation.
The complete implementation is shown below.

\begin{cardkey}[label={box:format_check}]
\small\ttfamily\obeylines\obeyspaces
def check\_format(text):
~~~~"""Format check: returns (True, "ok") if the output passes,
~~~~or (False, <label>) identifying the first violation found."""
~
~~~~if text == "":
~~~~~~~~return False, "empty"
~
~~~~if "\#\# " in text:
~~~~~~~~return False, "problem\_marker"
~
~~~~prefix = text.strip()[:60].lower()
~
~~~~if "rewrit" in prefix:
~~~~~~~~return False, "mention\_rewrite"
~
~~~~if "process" in prefix or "reasoning" in prefix:
~~~~~~~~return False, "mention\_reasoning"
~
~~~~if ("bridge" in prefix or "clari" in prefix
~~~~~~~~~~~~or "deriv" in prefix or "refin" in prefix):
~~~~~~~~return False, "mention\_meta"
~
~~~~return True, "ok"
\end{cardkey}

\subsubsection{FSF Judge}
\label{sec:appendix_fsf}

We compute FSF using an LLM-as-a-Judge approach following~\citet{DBLP:journals/corr/abs-2509-19284},
which parses each rationale into a Graphviz DOT graph and defines FSF as the
fraction of failed nodes among all nodes.
Because querying a large judge model at every GRPO rollout is prohibitively expensive,
we distill this judgment procedure into a lightweight judge via supervised fine-tuning.
Specifically, we sample rationales from OmniThought~\cite{DBLP:journals/corr/abs-2505-10937}%
\footnote{\url{https://huggingface.co/datasets/alibaba-pai/OmniThought}},
annotate them with GPT-OSS-120B~\cite{DBLP:journals/corr/abs-2508-10925} to construct
49,000 DOT-labeled examples, and fine-tune Qwen3-4B on the resulting FSF judge
training set.\footnote{
The FSF judge training set is available at
\url{https://huggingface.co/datasets/AnonymousCard/fsf_grm_train}.
}
The fine-tuning configuration is summarized below.
The prompt template used for both teacher annotation and judge inference is shown afterward.

\begin{cardbox}{{\small\faCogs}\; Fine-tuning Configuration for the Lightweight FSF Judge}
\begin{lstlisting}[style=cardtree]
fsf_judge:
  base_model: Qwen3-4B
  objective: DOT-graph failure-node judgment

training:
  learning_rate: 2e-05
  num_epochs: 2.0
  seed: 42

batching:
  train_batch_size_per_device: 2
  eval_batch_size_per_device: 1
  num_devices: 4
  total_train_batch_size: 8
  total_eval_batch_size: 4
  distributed_type: multi-GPU

optimization:
  optimizer: AdamW
  betas: [0.9, 0.999]
  epsilon: 1e-08
  lr_scheduler_type: cosine
  lr_scheduler_warmup_ratio: 0.1

framework_versions:
  transformers: 4.55.0
  pytorch: 2.8.0+cu128
  datasets: 3.6.0
  tokenizers: 0.21.4
\end{lstlisting}
\end{cardbox}

\begin{cardprompt}[label={box:fsf_prompt}]{FSF Judge Prompt Template}
\small\ttfamily\obeylines\obeyspaces
Parse the reasoning trajectory into a Graphviz diagram. Focus on these essentials.
~
Node Rules
- One node per distinct reasoning step.
- fillcolor=lightblue: Successful reasoning steps.
- fillcolor=lightpink: Failed attempts.
~
Edge Rules
- Connect node A -> node B if the information or insight from A is
~~actually used to construct the reasoning in B.
- Branch new attempts from their starting ancestor, not from dead ends.
~
Requirements
- Use rankdir=TB.
- Include all attempts, including failures.
- Always start with a problem statement node.
- Always end with a final answer node.
- Do not reorder or reorganize the reasoning flow.
~
Output Format
- Generate complete Graphviz DOT code in dot blocks.
\end{cardprompt}

\subsection{Training Environment and Code Structure}
\label{sec:appendix_codebase}
 
CARD is implemented in Python~3.12 with PyTorch~2.6.0 and CUDA~12.4,
and builds on the VERL reinforcement learning framework~\footnote{\url{https://github.com/verl-project/verl}}.
The codebase is organized around two main stages.
The first stage performs \emph{CoT rewriting}: a Reformatter is trained with
GRPO using the distribution alignment and logical coherence rewards, then rewrites
teacher-generated reasoning chains into training trajectories.
The second stage performs \emph{supervised fine-tuning (SFT)}, where the student
model is trained on the reformatted trajectories.
 
\vspace{0.4em}
 
\begin{cardbox}{{\small\faFolder}\; Codebase Structure}
\begin{lstlisting}[style=cardtree]
CARD/
(*\textcolor{cardBlue}{+--}*) eval/                        (*\textcolor{cardSlate}{\# Evaluation toolkit}*)
(*\textcolor{cardBlue}{|}*)   (*\textcolor{cardBlue}{+--}*) eval.py               (*\textcolor{cardSlate}{\# Cross-benchmark streaming eval engine}*)
(*\textcolor{cardBlue}{|}*)   (*\textcolor{cardBlue}{+--}*) collect_results.py       (*\textcolor{cardSlate}{\# Result aggregation across checkpoints}*)
(*\textcolor{cardBlue}{|}*)   (*\textcolor{cardBlue}{+--}*) eval_configs/            (*\textcolor{cardSlate}{\# Per-benchmark YAML configs}*)
(*\textcolor{cardBlue}{|}*)   (*\textcolor{cardBlue}{+--}*) launch_scripts/          (*\textcolor{cardSlate}{\# Batch evaluation shell scripts}*)
(*\textcolor{cardBlue}{+--}*) cot_rewriter_grpo/          (*\textcolor{cardSlate}{\# Core training \& inference}*)
(*\textcolor{cardBlue}{|}*)   (*\textcolor{cardBlue}{+--}*) verl/                    (*\textcolor{cardSlate}{\# VERL framework (forked \& extended)}*)
(*\textcolor{cardBlue}{|}*)   (*\textcolor{cardBlue}{|}*)   (*\textcolor{cardBlue}{+--}*) verl/trainer/        (*\textcolor{cardSlate}{\# PPO / GRPO trainer entry points}*)
(*\textcolor{cardBlue}{|}*)   (*\textcolor{cardBlue}{|}*)   (*\textcolor{cardBlue}{+--}*) verl/workers/        (*\textcolor{cardSlate}{\# FSDP / vLLM / SGLang workers}*)
(*\textcolor{cardBlue}{|}*)   (*\textcolor{cardBlue}{|}*)   (*\textcolor{cardBlue}{+--}*) verl/experimental/   (*\textcolor{cardSlate}{\# Custom CoT-rewriter reward functions}*)
(*\textcolor{cardBlue}{|}*)   (*\textcolor{cardBlue}{+--}*) inference/               (*\textcolor{cardSlate}{\# Rollout \& dataset construction pipeline}*)
(*\textcolor{cardBlue}{|}*)   (*\textcolor{cardBlue}{|}*)   (*\textcolor{cardBlue}{+--}*) pipeline.py          (*\textcolor{cardSlate}{\# Steps 1--4: load, rollout, judge, FSF tag}*)
(*\textcolor{cardBlue}{|}*)   (*\textcolor{cardBlue}{|}*)   (*\textcolor{cardBlue}{+--}*) build_sharegpt.py    (*\textcolor{cardSlate}{\# Convert rollouts to ShareGPT format}*)
(*\textcolor{cardBlue}{|}*)   (*\textcolor{cardBlue}{|}*)   (*\textcolor{cardBlue}{+--}*) configs/             (*\textcolor{cardSlate}{\# Inference YAML configs per experiment}*)
(*\textcolor{cardBlue}{|}*)   (*\textcolor{cardBlue}{+--}*) scripts/                 (*\textcolor{cardSlate}{\# Training launch scripts}*)
(*\textcolor{cardBlue}{|}*)   (*\textcolor{cardBlue}{|}*)   (*\textcolor{cardBlue}{+--}*) grpo/run_train.sh    (*\textcolor{cardSlate}{\# Main GRPO training script (Hydra)}*)
(*\textcolor{cardBlue}{|}*)   (*\textcolor{cardBlue}{|}*)   (*\textcolor{cardBlue}{+--}*) setup_env.sh         (*\textcolor{cardSlate}{\# Environment variable setup}*)
(*\textcolor{cardBlue}{|}*)   (*\textcolor{cardBlue}{+--}*) utils/                   (*\textcolor{cardSlate}{\# Model preparation \& tokenizer utilities}*)
(*\textcolor{cardBlue}{|}*)   (*\textcolor{cardBlue}{+--}*) examples/                (*\textcolor{cardSlate}{\# Training config YAMLs (per model / GPU)}*)
(*\textcolor{cardBlue}{+--}*) requirements.txt             (*\textcolor{cardSlate}{\# Unified dependency list}*)
\end{lstlisting}
\end{cardbox}
 
\vspace{0.5em}


\section{Benchmarks and Baselines}
\label{sec:appendix_benchmarks}

\subsection{Benchmark Descriptions}
\label{sec:appendix_benchmark_desc}

\paragraph{In-Domain Benchmarks.}
\begin{description}
  \item[\textbf{MATH-500}~\cite{DBLP:conf/iclr/LightmanKBEBLLS24}]
    is a 500-problem representative subset of the MATH benchmark, spanning
    seven mathematical subjects (prealgebra, algebra, number theory, counting
    and probability, geometry, intermediate algebra, and precalculus) at
    competition difficulty; we report accuracy using exact-match answer
    normalization.

  \item[\textbf{GSM8K}~\cite{DBLP:journals/corr/abs-2110-14168}]
    comprises 8{,}500 grade-school mathematics problems requiring multi-step
    arithmetic reasoning and basic algebra, and serves as a standard probe
    for elementary mathematical reasoning.

  \item[\textbf{AGI-MATH}~\cite{DBLP:conf/naacl/ZhongCGLLWSCD24}]
    is the mathematical subset of the AGIEval benchmark, which sources
    problems from human-centric standardized exams; the math-relevant tasks
    include SAT Math, AQuA-RAT quantitative reasoning, and Gaokao math in
    both multiple-choice and cloze formats.

    \item[\textbf{AIME 2024} and \textbf{AIME 2025}]
AIME 2024\footnote{\url{https://modelscope.cn/datasets/evalscope/aime24}}
and AIME 2025\footnote{\url{https://modelscope.cn/datasets/evalscope/aime25}}
each consist of 30 problems from the American Invitational Mathematics
Examination in the corresponding year. These benchmarks evaluate performance
on high-difficulty mathematical reasoning tasks, where models are required to
carry out sustained multi-step problem solving.

  \item[\textbf{AMC 2023}]
AMC (American Mathematics Competitions) is a benchmark based on problems from the AMC 10/12 competitions from 2023. These multiple-choice problems test mathematical problem-solving skills at the high school level and serve as qualifiers for the AIME competition.~\footnote{\url{https://modelscope.cn/datasets/evalscope/amc_22-24}}
\end{description}

\paragraph{Out-of-Domain Benchmarks.}
\begin{description}
    \item[\textbf{MMLU-Pro (No Math)}~\cite{DBLP:conf/nips/WangMZNCGRAHJLK24}]
    builds on MMLU~\cite{DBLP:conf/iclr/HendrycksBBZMSS21}, a widely used benchmark
    that covers 57 academic subjects with four-choice questions.
    It introduces more reasoning-focused questions and expands each question
    to ten answer choices, which substantially raises difficulty over the
    original format. We exclude all mathematics subcategories to isolate
    generalization to non-mathematical knowledge and reasoning.

  \item[\textbf{ARC-Challenge}~\cite{DBLP:journals/corr/abs-1803-05457}]
    is the challenging split of the AI2 Reasoning Challenge, comprising
    2{,}590 natural science questions that both retrieval-based and
    co-occurrence methods fail to answer correctly; we use it as a
    lightweight probe for catastrophic forgetting of general scientific
    reasoning induced by math-focused SFT.
\end{description}

\subsection{Baseline Descriptions}
\label{sec:appendix_baselines}

All baselines are constructed from the same candidate trajectory pool and matched to a 30,000-example training budget where applicable.
\begin{description}
  \item[\textbf{Vanilla SFT}]
    trains the student directly on raw DeepSeek-R1-671B solver trajectories
    without any filtering or reformatting, establishing the performance
    achievable by vanilla \emph{trajectory-based} knowledge distillation
    under the same data budget.

  \item[\textbf{MCC-KD}~\cite{DBLP:conf/emnlp/ChenWQWYZ23}]
    generates multiple \emph{rationales} per question and retains those that
    maximize cross-\emph{rationale} answer consistency by minimizing the
    bidirectional KL divergence between the induced answer distributions.

  \item[\textbf{MoRSD}~\cite{DBLP:conf/emnlp/YanLPCXT25}]
    computes a \emph{Rationale Difficulty} score for each trajectory,
    measured by the student's perplexity on the correct answer given the
    \emph{rationale}, and selects a compact subset that jointly balances
    accuracy, diversity, and difficulty.

\item[\textbf{Mix Small Teacher}]~\cite{DBLP:journals/corr/abs-2501-12948}
    replaces the 8{,}819 trajectories whose \emph{rationales} satisfy
    $\mathrm{FSF} > 0$ with trajectories sampled from a small teacher model,
    using the same model that initializes the CARD Reformatter before training.
    Specifically, we use Llama-3.1-8B-Instruct as the small teacher for
    Llama student models, and Qwen2.5-7B-Instruct for the Qwen2.5-7B-Instruct
    student (i.e., the student itself serves as the small teacher).
    For each such question, we sample from the small teacher and retain only
    trajectories that yield a correct answer, substituting them into the
    30{,}000-example training set.
    Sampling uses temperature $0.6$, top-$p$ $0.95$, top-$k$ $25$,
    a maximum of $28{,}000$ tokens, and no few-shot examples.
    
  \item[\textbf{Mix TokenSkip}]~\cite{DBLP:conf/emnlp/XiaLWLL25}
    applies the same substitution logic as Mix Small Teacher, but replaces
    the \emph{rationales} of the $\mathrm{FSF} > 0$ subset using
    TokenSkip, which removes redundant low-importance
    tokens from the \emph{rationale} while preserving its structure.
    Token importance scores are computed with LLMLingua-2~\footnote{\url{https://huggingface.co/microsoft/llmlingua-2-xlm-roberta-large-meetingbank}} 
    with a target keep rate of $0.7$, retaining the $70\%$ most important
    \emph{rationale} tokens.
\end{description}


\section{Experimental Details}
\label{sec:appendix_training}

This section summarizes the implementation details of CARD training. We separate
the experimental setup into three parts: GRPO training for the CARD Reformatter,
supervised fine-tuning for knowledge distillation, and the computational cost
of the full pipeline.

\subsection{CARD Reformatter Training via GRPO}
\label{sec:appendix_grpo}

The CARD Reformatter is optimized with GRPO on rewriting instances. The training
configuration includes the rollout setup, actor adaptation strategy, reward
composition, and FSF judge inference settings, as summarized below.

\begin{cardbox}{{\small\faCogs}\; GRPO Configuration for the CARD Reformatter}
\begin{lstlisting}[style=cardtree]
grpo_training:
  num_rewriting_instances: 5000
  num_optimization_steps: 70
  global_batch_size: 64
  group_size: 8

sequence_lengths:
  max_prompt_length: 9000
  max_response_length: 8192

actor_lora:
  enabled: true
  rank: 64
  alpha: 128
  dropout: 0.05
  target_modules: all-linear
  learning_rate: 5e-05

reward:
  rationale_alignment_weight: 0.75
  answer_alignment_weight: 0.25

fsf_judge_inference:
  temperature: 0.4
  top_k: 20
  top_p: 0.95
  max_new_tokens: 3000
  context_window: 12000
\end{lstlisting}
\end{cardbox}

\subsection{Knowledge Distillation via SFT}
\label{sec:appendix_sft}

We warm-start the student models through supervised fine-tuning. The setup uses
different adaptation strategies for smaller and larger student models while
keeping the main training schedule consistent across settings.

\begin{cardbox}{{\small\faGraduationCap}\; SFT Configuration for Knowledge Distillation}
\begin{lstlisting}[style=cardtree]
sft_common:
  framework: LlamaFactory
  num_epochs: 2
  lr_scheduler_type: cosine
  warmup_ratio: 0.1
  batch_size: 16
  max_sequence_length: 9000
  precision: bf16

models_up_to_3b:
  tuning_method: full fine-tuning
  learning_rate: 2e-05

models_7b:
  tuning_method: LoRA
  learning_rate: 2e-04
  lora_rank: 128
  lora_alpha: 256
  lora_dropout: 0.05
  target_modules: all-linear
\end{lstlisting}
\end{cardbox}

\subsection{Computational Cost}
\label{app:compute}

We report the hardware setup, training frameworks, stage-wise runtime, and resulting GPU-hour estimate for the full CARD pipeline under the Qwen2.5-7B student setting. GPU-hours are computed as the product of the number of GPUs and wall-clock runtime. The cost summary below corresponds to this Qwen2.5-7B setting.

\begin{cardbox}{{\small\faServer}\; Computational Cost Summary}
\begin{lstlisting}[style=cardtree]
hardware:
  node_count: 1
  gpu_type: NVIDIA A100-PCIE-40GB
  num_gpus: 8

software:
  grpo_framework: VERL
  sft_framework: LlamaFactory

runtime_qwen2_5_7b:
  lora_sft_hours: 4
  card_optimization_hours: 36
  benchmark_evaluation_hours: 12
  total_wall_clock_hours: 52

gpu_hours:
  formula: num_gpus * total_wall_clock_hours
  calculation: 8 * 52
  total_a100_gpu_hours: 416
\end{lstlisting}
\end{cardbox}
\subsection{Full Experiment Results}
\label{sec:appendix_ablation_full}
\label{sec:appendix_fsf_controls}
\label{sec:appendix_seed_evaluation}

This section presents the complete per-benchmark results for FSF filtering controls (Section~\ref{sec:fsf_filtering}, Table~\ref{tab:fsf_filtering_full}), random-seed evaluation (Section~\ref{sec:seed_robustness}, Table~\ref{tab:seed_results_full}), cross-model transferability (Section~\ref{sec:card_transferability}, Table~\ref{tab:card_transferability_full}), CARD component ablations (Section~\ref{sec:ablation_components}, Table~\ref{tab:ablation_components_full}), and FSF judge robustness (Section~\ref{sec:fsf_judge_robustness}, Table~\ref{tab:fsf_judge_robustness_full}).

\providecommand{\abest}[1]{\cellcolor{bestcolor}\textbf{#1}}
\providecommand{\asecond}[1]{\cellcolor{secondcolor}\uline{#1}}

\begingroup
\makeatletter
\setlength{\@fptop}{0pt}
\setlength{\@fpsep}{18pt}
\setlength{\@fpbot}{0pt plus 1fil}
\makeatother
\newcommand{\avgcell}[2]{%
  \makebox[2.4em][r]{#1}%
  \makebox[2em][l]{#2}%
}

\begin{table}[htbp]
\centering
\caption{Full FSF filtering comparison on Qwen2.5-7B. Math Avg and Gen Avg
are macro-averages over the six math and two OOD benchmarks, respectively. $\uparrow/\downarrow$ denote changes from Vanilla SFT.}
\label{tab:fsf_filtering_full}
\setlength{\tabcolsep}{3pt}
\renewcommand{\arraystretch}{1.18}
\footnotesize
\resizebox{\linewidth}{!}{%
\begin{tabular}{@{}lccccccccccc}
\toprule
\textbf{Method} & \textbf{\#Train} & \textbf{AIME24} & \textbf{AIME25} & \textbf{GSM8K} & \textbf{MATH500} & \textbf{AMC23} & \textbf{AGI-Math} & \textbf{Math Avg} & \textbf{MMLU-Pro} & \textbf{ARC-C} & \textbf{Gen Avg} \\
\midrule
Vanilla SFT & 30,000 & 13.89 & 11.25 & 90.58 & 75.80 & 45.40 & 73.90 & \avgcell{51.80}{} & 45.60 & \asecond{67.07} & \avgcell{56.34}{} \\
Drop-FSF & 21,181 & 13.16 & 10.40 & 90.20 & 74.30 & 44.49 & 73.03 & \avgcell{50.93}{\,\down{0.87}} & 52.90 & 59.98 & \avgcell{56.44}{\,\up{0.10}} \\
Replace-FSF & 30,000 & \asecond{14.87} & \asecond{12.39} & \asecond{91.09} & \asecond{76.42} & \asecond{46.61} & \asecond{75.06} & \cellcolor{secondcolor}\avgcell{\uline{52.74}}{\,\up{0.94}} & \asecond{53.50} & 61.60 & \cellcolor{secondcolor}\avgcell{\uline{57.55}}{\,\up{1.21}} \\
CARD & 30,000 & \abest{16.39} & \abest{14.16} & \abest{91.87} & \abest{77.50} & \abest{48.50} & \abest{76.87} & \cellcolor{bestcolor}\avgcell{\textbf{54.22}}{\,\up{2.42}} & \abest{54.00} & \abest{88.14} & \cellcolor{bestcolor}\avgcell{\textbf{71.07}}{\,\up{14.73}} \\
\bottomrule
\end{tabular}}
\end{table}

\begin{table}[htbp]
\centering
\caption{Full per-seed comparison of Vanilla SFT and CARD on Qwen2.5-7B across three training seeds. $\uparrow$ denotes improvement over Vanilla SFT with the same seed.}
\label{tab:seed_results_full}
\footnotesize
\setlength{\tabcolsep}{3pt}
\renewcommand{\arraystretch}{1.18}
\resizebox{\linewidth}{!}{%
\begin{tabular}{@{}lccccccccccc}
\toprule
\textbf{Method} & \textbf{Seed} & \textbf{AIME24} & \textbf{AIME25} & \textbf{GSM8K} & \textbf{MATH500} & \textbf{AMC23} & \textbf{AGI-Math} & \textbf{Math Avg} & \textbf{MMLU-Pro} & \textbf{ARC-C} & \textbf{Gen Avg} \\
\midrule
Vanilla SFT & 42 & \asecond{13.89} & \asecond{11.25} & \asecond{90.58} & \asecond{75.80} & \asecond{45.40} & \asecond{73.90} & \cellcolor{secondcolor}\avgcell{\uline{51.80}}{} & \asecond{45.60} & \asecond{67.07} & \cellcolor{secondcolor}\avgcell{\uline{56.34}}{} \\
CARD & 42 & \abest{16.39} & \abest{14.16} & \abest{91.87} & \abest{77.50} & \abest{48.50} & \abest{76.87} & \cellcolor{bestcolor}\avgcell{\textbf{54.22}}{\,\up{2.42}} & \abest{54.00} & \abest{88.14} & \cellcolor{bestcolor}\avgcell{\textbf{71.07}}{\,\up{14.73}} \\
\midrule
Vanilla SFT & 307 & \asecond{15.70} & \asecond{13.17} & \asecond{90.25} & \asecond{75.36} & \asecond{48.47} & \asecond{75.27} & \cellcolor{secondcolor}\avgcell{\uline{53.04}}{} & \asecond{45.98} & \asecond{67.33} & \cellcolor{secondcolor}\avgcell{\uline{56.66}}{} \\
CARD & 307 & \abest{17.28} & \abest{17.70} & \abest{92.23} & \abest{79.51} & \abest{48.90} & \abest{77.90} & \cellcolor{bestcolor}\avgcell{\textbf{55.59}}{\,\up{2.55}} & \abest{53.38} & \abest{86.10} & \cellcolor{bestcolor}\avgcell{\textbf{69.74}}{\,\up{13.08}} \\
\midrule
Vanilla SFT & 1234 & \asecond{14.15} & \asecond{13.62} & \asecond{90.06} & \asecond{77.19} & \asecond{43.51} & \asecond{74.51} & \cellcolor{secondcolor}\avgcell{\uline{52.17}}{} & \asecond{44.85} & \asecond{68.86} & \cellcolor{secondcolor}\avgcell{\uline{56.86}}{} \\
CARD & 1234 & \abest{17.28} & \abest{16.31} & \abest{91.65} & \abest{78.11} & \abest{50.78} & \abest{75.93} & \cellcolor{bestcolor}\avgcell{\textbf{55.01}}{\,\up{2.84}} & \abest{54.22} & \abest{88.84} & \cellcolor{bestcolor}\avgcell{\textbf{71.53}}{\,\up{14.67}} \\
\bottomrule
\end{tabular}}
\end{table}

\begin{table*}[h]
\centering
\small
\setlength{\tabcolsep}{3.2pt}
\caption{Full cross-model transferability of the CARD Reformatter on Llama3.2-3B. \textit{CARD Train By} denotes the student model used to compute rewards during Reformatter training.}
\label{tab:card_transferability_full}
\renewcommand{\arraystretch}{1.18}
\resizebox{\textwidth}{!}{
\begin{tabular}{lcccccccccc}
\toprule
\textbf{Reward Student} &
\textbf{AIME24} &
\textbf{AIME25} &
\textbf{GSM8K} &
\textbf{MATH500} &
\textbf{AMC23} &
\textbf{AGI-MATH} &
\textbf{Math Avg} &
\textbf{\makecell{MMLU-Pro}} &
\textbf{ARC-C} &
\textbf{Gen Avg} \\
\midrule
Llama3.2-1B
& 2.64 & 0.28 & 76.23 & 46.40 & 21.33 & 46.68 & 32.26 & 36.02 & 74.23 & 55.13 \\
\cellcolor{bestcolor}\textbf{Llama3.2-3B}
& \cellcolor{bestcolor}\textbf{3.74}
& \cellcolor{bestcolor}\textbf{1.38}
& \cellcolor{bestcolor}\textbf{77.26}
& \cellcolor{bestcolor}\textbf{49.40}
& \cellcolor{bestcolor}\textbf{23.13}
& \cellcolor{bestcolor}\textbf{48.60}
& \cellcolor{bestcolor}\textbf{33.92}
& \cellcolor{bestcolor}\textbf{36.91}
& \cellcolor{bestcolor}\textbf{74.74}
& \cellcolor{bestcolor}\textbf{55.83} \\
Qwen2.5-7B
& 2.78 & 0.56 & 77.05 & 46.15 & 23.37 & 48.22 & 33.02 & \textbf{37.23} & 73.24 & 55.24 \\
\bottomrule
\end{tabular}
}
\end{table*}

\begin{table*}[h]
\centering
\small
\setlength{\tabcolsep}{3.2pt}
\caption{Full ablation on CARD components using Qwen2.5-7B with LoRA. $\downarrow$ denotes the drop compared with full CARD.}
\label{tab:ablation_components_full}
\resizebox{\textwidth}{!}{
\begin{tabular}{lcccccccccc}
\toprule
\textbf{Method}
& \textbf{AIME24}
& \textbf{AIME25}
& \textbf{GSM8K}
& \textbf{MATH500}
& \textbf{AMC23}
& \textbf{AGI-MATH}
& \textbf{Math Avg}
& \textbf{\makecell{MMLU-Pro}}
& \textbf{ARC-C}
& \textbf{Gen Avg} \\
\midrule
CARD
& \abest{16.39} & \abest{14.16} & \abest{91.87} & \abest{77.50}
& \abest{48.50} & \asecond{76.87}
& \abest{54.22}
& \asecond{54.00} & \abest{88.14}
& \abest{71.07} \\
CARD w/o $r_{\mathrm{coh}}$
& 15.00 & \asecond{13.61} & \asecond{90.07} & 74.57
& 46.46 & 75.50
& 52.54\,\down{1.68}
& 52.82 & 79.80
& 66.31\,\down{4.76} \\
CARD w/o $r_{\mathrm{rat}}$
& \asecond{15.13} & 12.36 & 89.50 & 73.42
& 46.46 & 76.50
& 52.23\,\down{1.99}
& \abest{54.14} & 82.50
& 68.32\,\down{2.75} \\
CARD w/o $r_{\mathrm{ans}}$
& 14.02 & 12.36 & 89.61 & 73.90
& \asecond{46.60} & 75.90
& 52.07\,\down{2.15}
& 53.90 & 83.35
& 68.63\,\down{2.44} \\
CARD w/o $\mathbb{I}_{\mathrm{leak}}$
& 14.86 & 13.19 & 89.54 & \asecond{74.77}
& \asecond{46.60} & \abest{77.00}
& \asecond{52.66}\,\down{1.56}
& 52.81 & \asecond{85.52}
& \asecond{69.17}\,\down{1.90} \\
\bottomrule
\end{tabular}
}
\end{table*}

\begin{table*}[h]
\centering
\small
\setlength{\tabcolsep}{3.2pt}
\caption{Full Robustness to FSF judge accuracy using Qwen2.5-7B with LoRA. $\downarrow$ denotes the drop compared with full CARD.}
\label{tab:fsf_judge_robustness_full}
\resizebox{\textwidth}{!}{
\begin{tabular}{lcccccccccc}
\toprule
\textbf{Method}
& \textbf{AIME24}
& \textbf{AIME25}
& \textbf{GSM8K}
& \textbf{MATH500}
& \textbf{AMC23}
& \textbf{AGI-MATH}
& \textbf{Math Avg}
& \textbf{\makecell{MMLU-Pro \\ (w/o math)}}
& \textbf{ARC-C}
& \textbf{Gen Avg} \\
\midrule
CARD
& \abest{16.39} & \abest{14.16} & \abest{91.87} & \asecond{77.50}
& \abest{48.50} & \asecond{76.87}
& \abest{54.22}
& 54.00 & \abest{88.14}
& \abest{71.07} \\
CARD w/ 5\% judge bias
& 13.05 & \asecond{13.88} & 89.78 & 76.60
& \asecond{47.83} & 76.15
& 52.88\,\down{1.34}
& \abest{55.22} & 82.17
& 68.70\,\down{2.37} \\
CARD w/ 10\% judge bias
& \asecond{14.86} & 12.78 & \asecond{90.47} & \abest{77.58}
& 47.28 & \abest{77.27}
& \asecond{53.37}\,\down{0.85}
& \asecond{53.13} & \asecond{84.75}
& \asecond{68.94}\,\down{2.13} \\
CARD w/ 20\% judge bias
& 14.72 & 13.05 & 90.33 & 75.40
& 47.14 & 76.80
& 52.91\,\down{1.31}
& 51.78 & 81.06
& 66.42\,\down{4.65} \\
\bottomrule
\end{tabular}
}
\end{table*}

\FloatBarrier
\endgroup
\newpage

\begingroup
\raggedbottom
\setlength{\abovedisplayskip}{3pt}
\setlength{\belowdisplayskip}{3pt}
\setlength{\abovedisplayshortskip}{2pt}
\setlength{\belowdisplayshortskip}{2pt}
\section{Case Study}
\label{sec:appendix_case_study}

This section presents a case study showing how CARD reformatter transforms the original rationale of a given question into a cognition-aligned rationale. For readability, we keep only the beginning, ending, and key intermediate parts of the original rationale when it is long.

\begin{cardprompt}{Question}
\small
The expression $3y^2-y-24$ can be written as $(3y + a)(y + b)$, where $a$ and $b$ are integers. What is $a - b$?
\end{cardprompt}

\noindent
\begin{minipage}[t]{0.49\linewidth}
\vspace{0pt}
\begin{cardorig}[left=2mm,right=2mm,top=2mm,bottom=2mm,before skip=4pt,after skip=0pt]{Original Rationale \normalfont\small (excerpt)}
\small
Okay, so I have this algebra problem here: I need to factor the quadratic expression $3y^2-y-24$ into the form $(3y+a)(y+b)$, where $a$ and $b$ are integers. Then, I have to find the value of $a-b$. Since the problem already tells us the form it should take, which is $(3y+a)(y+b)$, that might help.

First, let me recall that when you multiply two binomials, you use the FOIL method. If I expand $(3y+a)(y+b)$, I get
\[
\begin{aligned}
&(3y+a)(y+b)\\
&\quad=3y^2+3by+ay+ab\\
&\quad=3y^2+(3b+a)y+ab.
\end{aligned}
\]
Now, the original quadratic is $3y^2-y-24$. Comparing coefficients, we can set up equations:
\[
3b+a=-1,\qquad ab=-24.
\]

\smallskip
\noindent
$\cdots$

\smallskip
\noindent
Solving the resulting equations gives $b=-3$ and $a=8$.

\smallskip
\noindent
$\cdots$

\smallskip
\noindent
Multiply it out:
\[
\begin{aligned}
3y\cdot y&=3y^2,\quad 3y\cdot(-3)=-9y,\\
8\cdot y&=8y,\quad 8\cdot(-3)=-24.
\end{aligned}
\]
Combine:
\[
3y^2-9y+8y-24=3y^2-y-24.
\]
Yes, that's exactly the original expression. So, correct.

Therefore, the values of $a$ and $b$ are correct. Hence,
\[
a-b=8-(-3)=11.
\]
So, the answer is $11$.

\[
\text{Final Answer: }\boxed{11}
\]
\end{cardorig}
\end{minipage}\hfill
\begin{minipage}[t]{0.49\linewidth}
\vspace{0pt}
\begin{cardbox}[left=2mm,right=2mm,top=2mm,bottom=2mm,before skip=4pt,after skip=0pt]{Reformatted Rationale}
\small
To factor the quadratic expression $3y^2-y-24$ into the form $(3y+a)(y+b)$, where $a$ and $b$ are integers, we need to find $a$ and $b$ such that the expanded form matches the original expression.

First, expand $(3y+a)(y+b)$:
\[
\begin{aligned}
&(3y+a)(y+b)\\
&\quad=3y^2+(3b+a)y+ab.
\end{aligned}
\]
Comparing this with $3y^2-y-24$, we obtain
\[
3b+a=-1,\qquad ab=-24.
\]

We need to solve these equations for integers $a$ and $b$. From $ab=-24$, list the integer factor pairs and check which one satisfies $3b+a=-1$:
\[
\begin{aligned}
(1,-24):&\quad 3(-24)+1=-71,\\
(2,-12):&\quad 3(-12)+2=-34,\\
(3,-8):&\quad 3(-8)+3=-21,\\
(4,-6):&\quad 3(-6)+4=-14,\\
(6,-4):&\quad 3(-4)+6=-6,\\
(8,-3):&\quad 3(-3)+8=-1.
\end{aligned}
\]
Thus, the only pair that works is $(a,b)=(8,-3)$.

Finally,
\[
a-b=8-(-3)=8+3=11.
\]
Therefore, the answer is
\[
\boxed{11}.
\]
\end{cardbox}
\end{minipage}
\par
\clearpage
\endgroup


\newpage

\section*{NeurIPS Paper Checklist}

\begin{enumerate}

\item {\bf Claims}
    \item[] Question: Do the main claims made in the abstract and introduction accurately reflect the paper's contributions and scope?
    \item[] Answer: \answerYes{}
    \item[] Justification: The abstract and Section~\ref{sec:introduction} state the paper's contributions and scope: CARD reformats solver trajectories into cognition-aligned teaching trajectories without teacher weights or repeated sampling. The claims are supported by the method in Section~\ref{sec:method} and the empirical results in Section~\ref{sec:experiment} and Appendix~\ref{sec:appendix_ablation_full}.

\item {\bf Limitations}
    \item[] Question: Does the paper discuss the limitations of the work performed by the authors?
    \item[] Answer: \answerYes{}
    \item[] Justification: Section~\ref{sec:conclusion} explicitly identifies two scope boundaries: (i) the experiments focus on mathematical reasoning, so the conclusions may not directly generalize to code, dialogue, or open-ended instruction following; and (ii) the evaluation is limited to student models up to 7B parameters, so scaling CARD to larger students may require additional compute and engineering effort. The paper also analyzes the robustness of CARD to FSF judge bias in Section~\ref{sec:fsf_judge_robustness}, while the computational cost of the representative Qwen2.5-7B setting is reported in Appendix~\ref{app:compute}.

\item {\bf Theory assumptions and proofs}
    \item[] Question: For each theoretical result, does the paper provide the full set of assumptions and a complete (and correct) proof?
    \item[] Answer: \answerNA{}
    \item[] Justification: The paper proposes an empirical distillation framework and optimization objective, but does not present formal theoretical results, theorems, or proofs.

    \item {\bf Experimental result reproducibility}
    \item[] Question: Does the paper fully disclose all the information needed to reproduce the main experimental results of the paper to the extent that it affects the main claims and/or conclusions of the paper (regardless of whether the code and data are provided or not)?
    \item[] Answer: \answerYes{}
    \item[] Justification: Section~\ref{sec:experiment} specifies the training data and splits, teacher trajectory pool, student models, baselines, and evaluation protocol across the math and out-of-domain benchmarks. Appendix~\ref{sec:appendix_reformatter} provides the rationale Reformatter prompt, Appendix~\ref{sec:appendix_fsf} documents the reward computation and FSF judging procedure, and Appendix~\ref{sec:appendix_codebase} describes the codebase structure. Appendix~\ref{sec:appendix_training} further reports the GRPO and SFT configurations, evaluation setup, computational cost, and checkpoint/output organization needed to reproduce the main pipeline. The anonymized supplementary release includes the CARD training framework code and an accompanying README with environment setup and running instructions.

\item {\bf Open access to data and code}
    \item[] Question: Does the paper provide open access to the data and code, with sufficient instructions to faithfully reproduce the main experimental results, as described in supplemental material?
    \item[] Answer: \answerYes{} 
    \item[] Justification: An anonymized version of the code is provided in the supplementary material, including CARD training scripts, configuration files, evaluation scripts, and an accompanying README with instructions for reproducing the main pipeline. All datasets and benchmarks used in this work are publicly available, including DeepMath-103K, OmniThought, GSM8K, MATH-500, AGI-MATH, AIME 2024/2025, AMC 2023, MMLU-Pro, and ARC-Challenge. Section~\ref{sec:experiment} specifies the training splits, sample sizes, baselines, and evaluation protocol, and Appendix~\ref{sec:appendix_training} reports the GRPO/SFT configurations and computational details. Model checkpoints are publicly released by their original providers, including DeepSeek-R1-671B, Llama-3.2, Qwen2.5, Qwen3-4B, and GPT-OSS-120B.

\item {\bf Experimental setting/details}
    \item[] Question: Does the paper specify all the training and test details (e.g., data splits, hyperparameters, how they were chosen, type of optimizer) necessary to understand the results?
    \item[] Answer: \answerYes{} 
    \item[] Justification: Section~\ref{sec:setup} specifies the training data, data splits, benchmark groups, student models, baselines, and evaluation protocols. Appendix~\ref{sec:appendix_training} further reports GRPO/SFT hyperparameters, LoRA settings, context lengths, precision, and implementation frameworks.

\item {\bf Experiment statistical significance}
    \item[] Question: Does the paper report error bars suitably and correctly defined or other appropriate information about the statistical significance of the experiments?
    \item[] Answer: \answerNo{} 
    \item[] Justification:  Due to the computational cost of training large language models, we performed
training of each of our models only once, in line with the prior work. Thus, we could not
report error bars for our main experiments.

\item {\bf Experiments compute resources}
    \item[] Question: For each experiment, does the paper provide sufficient information on the computer resources (type of compute workers, memory, time of execution) needed to reproduce the experiments?
    \item[] Answer: \answerYes{}
    \item[] Justification: Appendix~\ref{app:compute} reports the compute hardware, training frameworks, wall-clock runtime, and GPU-hour estimate for the representative Qwen2.5-7B student setting. Specifically, the reported configuration uses one node with 8 NVIDIA A100 PCIe 40GB GPUs, VERL for CARD/GRPO optimization, and LlamaFactory for student SFT. The appendix further reports the stage-wise runtime: 4 hours for LoRA SFT, 36 hours for CARD optimization, and 12 hours for benchmark evaluation, for a total of 52 wall-clock hours and 416 A100 GPU-hours.
    
\item {\bf Code of ethics}
    \item[] Question: Does the research conducted in the paper conform, in every respect, with the NeurIPS Code of Ethics \url{https://neurips.cc/public/EthicsGuidelines}?
    \item[] Answer: \answerYes{}
    \item[] Justification: The authors have reviewed the NeurIPS Code of Ethics and the research conforms to it in every respect. The work uses publicly released model checkpoints, trajectory datasets, and evaluation benchmarks in their intended research roles, involves no human subjects or personally identifiable information, and preserves anonymity in the submitted version, including the anonymized supplementary code release. The proposed CARD framework is evaluated as a reasoning-distillation method and does not introduce deployments or applications involving safety-critical decisions.

\item {\bf Broader impacts}
    \item[] Question: Does the paper discuss both potential positive societal impacts and negative societal impacts of the work performed?
    \item[] Answer: \answerYes{}
    \item[] Justification: On the positive side, CARD can reduce the cost of deploying reasoning-capable language models by enabling smaller student models to learn more effectively from teacher-generated solver trajectories, improving accessibility and energy efficiency. On the negative side, more efficient reasoning distillation could in principle make capable models easier to deploy in unintended settings, and the resulting students may inherit errors or biases from the teacher trajectories or automatic failed-step judgments. However, CARD operates on top of existing model checkpoints and trajectory datasets, and does not introduce or release a new high-risk pretrained model, so the marginal additional risk over existing distillation methods is limited.
    
\item {\bf Safeguards}
    \item[] Question: Does the paper describe safeguards that have been put in place for responsible release of data or models that have a high risk for misuse (e.g., pre-trained language models, image generators, or scraped datasets)?
    \item[] Answer: \answerNA{}
    \item[] Justification: The paper does not release new pre-trained language models, generative models, scraped datasets, or other assets with a high risk for misuse. It releases only the CARD training/evaluation code and a methodology that operates on top of already publicly released teacher, student, and judge-model checkpoints (DeepSeek-R1-671B, Llama-3.2, Qwen2.5, Qwen3-4B, and GPT-OSS-120B) and publicly released datasets or benchmarks (DeepMath-103K, OmniThought, GSM8K, MATH-500, AGI-MATH, AIME 2024/2025, AMC 2023, MMLU-Pro, and ARC-Challenge), which are governed by their respective providers' licenses and usage policies.

\item {\bf Licenses for existing assets}
    \item[] Question: Are the creators or original owners of assets (e.g., code, data, models), used in the paper, properly credited and are the license and terms of use explicitly mentioned and properly respected?
    \item[] Answer: \answerYes{}
    \item[] Justification: All existing assets used are properly credited at first mention: model checkpoints and APIs (DeepSeek-R1-671B, Llama-3.2, Qwen2.5, Qwen3-4B, GPT-OSS-120B), datasets and benchmarks (DeepMath-103K, OmniThought, GSM8K, MATH-500, AGI-MATH, AIME 2024/2025, AMC 2023, MMLU-Pro, and ARC-Challenge), and software libraries or frameworks (VERL, LlamaFactory, LLMLingua-2, PyTorch, Transformers, datasets, and tokenizers). All assets are used in their intended research roles under their published licenses and terms of use; no terms of use have been violated.

\item {\bf New assets}
    \item[] Question: Are new assets introduced in the paper well documented and is the documentation provided alongside the assets?
    \item[] Answer: \answerYes{}
    \item[] Justification: The new assets introduced in this work are the CARD training framework code and the FSF judge training set. The FSF judge training set is released in anonymized form via an accompanying artifact link (see Appendix~\ref{sec:appendix_fsf}), and is documented through its data source (OmniThought), annotation pipeline (GPT-OSS-120B), target DOT-graph format, and the fine-tuning configuration for the lightweight FSF judge. The CARD codebase is documented through: (1) a structural overview of the repository in Appendix~\ref{sec:appendix_codebase}; (2) full training and reproducibility details in Appendix~\ref{sec:appendix_training}, covering GRPO and SFT configurations, FSF-guided trajectory selection, the evaluation pipeline, computational cost, and checkpoint/output organization; and (3) an accompanying README with setup and usage instructions. The Reformatter prompt and reward design are further detailed in Appendix~\ref{sec:appendix_reformatter} and Appendix~\ref{sec:appendix_fsf}, respectively, to clarify how these components are integrated within the CARD pipeline.

\item {\bf Crowdsourcing and research with human subjects}
    \item[] Question: For crowdsourcing experiments and research with human subjects, does the paper include the full text of instructions given to participants and screenshots, if applicable, as well as details about compensation (if any)? 
    \item[] Answer: \answerNA{}
    \item[] Justification: The paper does not involve crowdsourcing experiments or research with human subjects. All evaluations are conducted automatically using publicly released benchmarks and rule-based judging or answer-extraction procedures where applicable, with no human annotators recruited for this work.

\item {\bf Institutional review board (IRB) approvals or equivalent for research with human subjects}
    \item[] Question: Does the paper describe potential risks incurred by study participants, whether such risks were disclosed to the subjects, and whether Institutional Review Board (IRB) approvals (or an equivalent approval/review based on the requirements of your country or institution) were obtained?
    \item[] Answer: \answerNA{}
    \item[] Justification: The paper does not involve crowdsourcing or research with human subjects, so IRB approval is not applicable.

\item {\bf Declaration of LLM usage}
    \item[] Question: Does the paper describe the usage of LLMs if it is an important, original, or non-standard component of the core methods in this research? Note that if the LLM is used only for writing, editing, or formatting purposes and does \emph{not} impact the core methodology, scientific rigor, or originality of the research, declaration is not required.
    \item[] Answer: \answerYes{} 
    \item[] Justification: LLMs are central to the experimental subjects and methodological components of this paper. They serve as the models being distilled and evaluated (the teacher and student models), and also instantiate the main modules of CARD (the CARD Reformatter and the lightweight FSF judge). These uses are part of the proposed research framework itself, rather than external assistance for writing, editing, or formatting. The relevant details are documented in the paper and appendix (teacher-generated trajectories and student distillation in Sections~\ref{sec:preliminaries}--\ref{sec:experiment}; the Reformatter prompt in Appendix~\ref{sec:appendix_reformatter}; the reward design and FSF-judge procedure in Appendix~\ref{sec:appendix_fsf}; and training configurations in Appendix~\ref{sec:appendix_training}). We do not use any additional undisclosed LLM assistance that affects the methodology, scientific claims, experimental results, figures, tables, or references.


\end{enumerate}

\end{document}